\documentclass[lettersize,journal]{IEEEtran}
\IEEEoverridecommandlockouts
\usepackage{amsmath, amsthm, amssymb, amsfonts, mathtools, breqn}
\usepackage{graphicx}
\usepackage{wrapfig} 
\usepackage{textcomp}
\usepackage{caption}
\usepackage{subcaption} 
\usepackage{hyperref}
\hypersetup{
    colorlinks=true,
    linkcolor=blue,
    filecolor=magenta,      
    urlcolor=cyan,
}
\usepackage{xcolor}

\usepackage{algorithm,algcompatible,algpseudocode}
\algblock{ParFor}{EndParFor}
\algnewcommand\algorithmicparfor{\textbf{parfor}}
\algnewcommand\algorithmicpardo{\textbf{do}}
\algnewcommand\algorithmicendparfor{\textbf{end\ parfor}}
\algrenewtext{ParFor}[1]{\algorithmicparfor\ #1\ \algorithmicpardo}
\algrenewtext{EndParFor}{\algorithmicendparfor}

\DeclareMathOperator*{\argmax}{argmax}

\DeclareMathOperator*{\E}{\mathbb{E}}
\newcommand{\norm}[1]{\left\lVert#1\right\rVert}
\newcommand{\abs}[1]{\left|#1\right|}
\newtheorem{prop}{Proposition}

\newcommand{\set}[1]{\left\{#1\right\}}
\newcommand{\brk}[1]{\left(#1\right)}
\newcommand{\bsq}[1]{\left[#1\right]}

\newcommand{\prob}[1]{\mathbb P\brk{#1}}
\newcommand{\R}{\mathbb{R}}

\newcommand{\PP}{\mathbb{P}}

\newcommand{\Z}{\mathbb{Z}}
\DeclareMathAlphabet{\mymathbb}{U}{BOONDOX-ds}{m}{n}

\begin{document}

\title{Decision-Making for Autonomous Vehicles with Interaction-Aware Behavioral Prediction and Social-Attention Neural Network\\
\thanks{$^{*}$Xiao Li, Kaiwen Liu, Anouck Girard, and Ilya Kolmanovsky are with the Department of Aerospace Engineering, University of Michigan, Ann Arbor, MI 48109, USA. {\tt\small \{hsiaoli, kwliu, anouck, ilya\}@umich.edu}}

\thanks{$^{\dagger}$H. Eric Tseng      Retired Senior Technical Leader, Ford Research and Advanced Engineering,             Chief Technologist, Excelled Tracer LLC.
{\tt\small hongtei.tseng@gmail.com}}

\thanks{This research was supported by the University of Michigan / Ford Motor Company Alliance program.}
}

\author{Xiao Li$^{*}$, Kaiwen Liu$^{*}$, H. Eric Tseng$^{\dagger}$, Anouck Girard$^{*}$, Ilya Kolmanovsky$^{*}$}
\maketitle

\begin{abstract}
Autonomous vehicles need to accomplish their tasks while interacting with human drivers in traffic. It is thus crucial to equip autonomous vehicles with artificial reasoning to better comprehend the intentions of the surrounding traffic, thereby facilitating the accomplishments of the tasks. In this work, we propose a behavioral model that encodes drivers' interacting intentions into latent social-psychological parameters. Leveraging a Bayesian filter, we develop a receding-horizon optimization-based controller for autonomous vehicle decision-making which accounts for the uncertainties in the interacting drivers' intentions. For online deployment, we design a neural network architecture based on the attention mechanism which imitates the behavioral model with online estimated parameter priors. We also propose a decision tree search algorithm to solve the decision-making problem online. The proposed behavioral model is then evaluated in terms of its capabilities for real-world trajectory prediction. We further conduct extensive evaluations of the proposed decision-making module, in forced highway merging scenarios, using both simulated environments and real-world traffic datasets. The results demonstrate that our algorithms can complete the forced merging tasks in various traffic conditions while ensuring driving safety.
\end{abstract}

\begin{IEEEkeywords}
Autonomous Vehicles, Interaction-Aware Driving, Imitation Learning, Neural Networks, Traffic Modeling
\end{IEEEkeywords}

\section{Introduction}\label{sec:intro}
\begin{figure}[!t]
    \centering
    \includegraphics[width=0.49\textwidth]{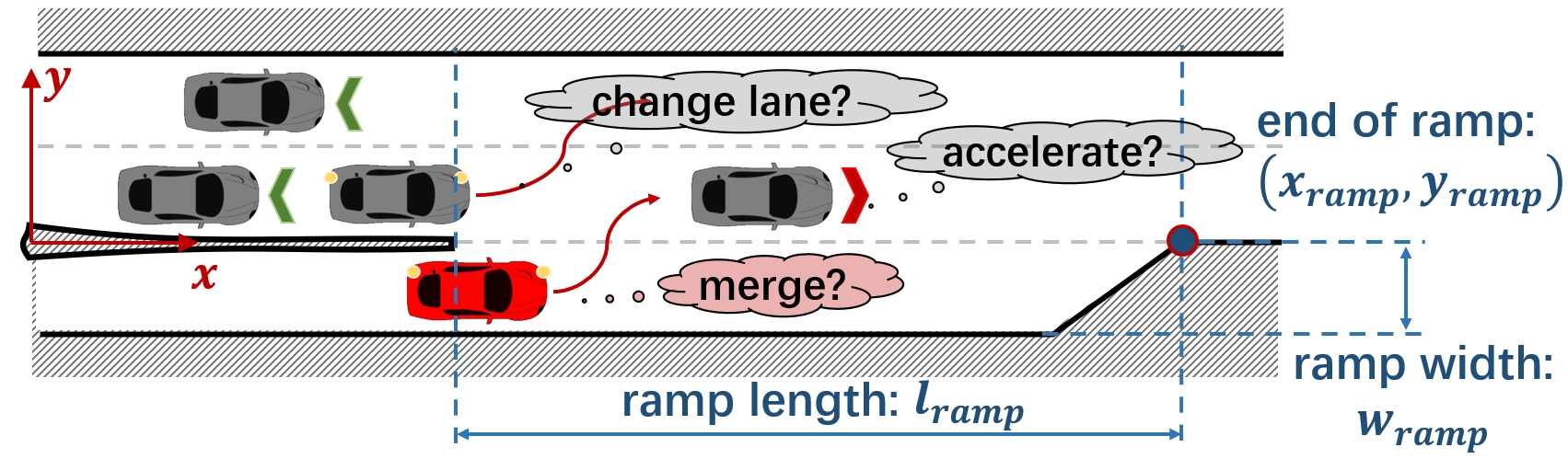}\vspace{-0.5em}
    \caption{Schematic diagram of the highway forced merging scenario: An on-ramp ego vehicle in red is merging onto the highway while interacting with the highway vehicles in grey.}
    \label{fig:forcedMergingProblem}
\end{figure}

One of the challenges in autonomous driving is interpreting the driving intentions of other human drivers. 
The communication between on-road participants is typically non-verbal, and relies heavily on turn/brake signals,
postures, eye contact, and subsequent behaviors. 
In uncontrolled traffic scenarios, e.g., roundabouts~\cite{polders2015identifying}, unsignalized intersections~\cite{haleem2010examining}, and highway ramps~\cite{mccartt2004types}, drivers need to negotiate their order of proceeding. 
Fig.~\ref{fig:forcedMergingProblem} illustrates a forced merging scenario at a highway entrance, where the ego vehicle in red attempts to merge into the highway before the end of the ramp. 
This merging action affects the vehicle behind in the lane being merged, 
and different social traits of its driver can result in different responses to the merging intent. A cooperative driver may choose a lane change to promote the merging process, while an egoistic driver may maintain a constant speed and disregard the merging vehicle. Therefore, understanding the latent intentions of other drivers can help the ego vehicle resolve conflicts and accomplish its task.

In this paper, we specifically focus on the highway forced merging scenario illustrated in Fig.~\ref{fig:forcedMergingProblem} and the objective of transitioning the ego vehicle onto the highway from the ramp in a timely and safe manner.  The difficulty of developing suitable automated driving algorithms for such scenarios is exacerbated by the fact that stopping on the ramp in non-congested highway traffic could be dangerous.   

%
%

The forced merging has been addressed in the automated driving literature from multiple directions.  In particular, learning-based methods have been investigated to synthesize controllers for such interactive scenarios. End-to-end planning methods~\cite{hu2023planning} have been proposed to generate control inputs from Lidar point clouds~\cite{caltagirone2017lidar} and RGB images~\cite{barnes2017find}. Reinforcement Learning (RL) algorithms have also been considered to learn end-to-end driving policies~\cite{hugle2020dynamic, mavrogiannis2022b}. 
A comprehensive survey of RL methods for autonomous driving applications is presented in~\cite{kiran2021deep}. 
Meanwhile, Imitation Learning-based methods have been 
exploited
to emulate expert driving behaviors~\cite{mei2021autonomous, pan2017agile}.
However, the end-to-end learning-based controllers lack interpretability and are limited in providing safety guarantees in unseen situations. To address these concerns, researchers have explored the integration of learning-based methods with planning and control techniques. 
Along these lines,
Model Predictive Control (MPC) algorithms have been integrated with the Social-GAN~\cite{gupta2018social} for trajectory prediction and planning~\cite{bae2022lane}. Meanwhile, Inverse RL methods have also been explored to predict drivers' behavior for planning purposes~\cite{sadigh2016planning, DanielaRusSVO, you2019advanced}. However, the learning-based modules in these systems may have limited capability to generalize and transfer to unobserved scenarios or behaviors.

There also exists extensive literature on modeling the interactive behaviors between drivers using model-based approaches. Assuming drivers maximize their rewards~\cite{sadigh2016planning},  game-theoretic approaches have been proposed to model traffic interactions, such as level-$k$ hierarchical reasoning framework~\cite{li2016hierarchical}, potential games~\cite{liu2022potential}, and Stackelberg games~\cite{DanielaRusSVO, fisac2019hierarchical}. In the setting of the level-$k$ game-theoretic models, approaches to estimating drivers' reasoning levels have been proposed~\cite{tian2021learning}. 
A novel Leader-Follower Game-theoretic Controller (LFGC) has been developed for decision-making in forced merging scenarios~\cite{liu2022interaction}. However, solving game-theoretic problems could be computationally demanding and has limited scalability to a larger number of interacting drivers or longer prediction horizons. 
To be able to account for the uncertainty in the interactions,
probabilistic methods, leveraging either Bayesian filter~\cite{liu2022interaction, wei2013autonomous} or particle filter~\cite{hoermann2017probabilistic} with Partially Observable Markov Decision Process (POMDP)~\cite{liu2022interaction, galceran2015multipolicy}, have also been implemented to encode and estimate the uncertain intent of other drivers as hidden variables.

In this paper, we consider the Social Value Orientation (SVO) from social psychology studies~\cite{messick1968motivational} to model drivers' interactions. The SVO quantifies subjects' tendencies toward social cooperation~\cite{liebrand1984effect} 
and has been previously used to model drivers' cooperativeness during autonomous vehicle decision-making in \cite{DanielaRusSVO, lenz2016tactical}. In addition, researchers have combined SVO-based rewards with RL to generate pro-social autonomous driving behaviors~\cite{crosato2021human, crosato2022interaction} or synthesize realistic traffic simulation with SVO agents~\cite{peng2021learning}. 
In our proposed behavioral model, we consider both social cooperativeness and the personal objectives of the interacting drivers. Leveraging a Bayesian filter, we propose a decision-making module that accounts for pairwise interactions with other drivers, and computes a reference trajectory for the ego vehicle under the uncertain cooperative intent of other drivers. The method proposed in this paper differs from the previous work~\cite{ACC2024} in the following aspects: 
1) Instead of using an action space with a few coarse action primitives, we synthesize a state-dependent set of smooth and realistic trajectories as our action space. 
2) We design a Social-Attention Neural Network (SANN) architecture to imitate the behavioral model that structurally incorporates the model-based priors and can be transferred to various traffic conditions. 
3) We develop a decision-tree search algorithm for the ego vehicle's decision-making, which guarantees safety and scalability. 
4) We conduct an extensive evaluation of the behavioral model in predicting real-world trajectory and demonstrate the decision-making module capabilities in forced merging scenarios on both simulations and real-world datasets, which is not done in~\cite{ACC2024}. 

The proposed algorithm has several distinguished features:
\begin{enumerate}
    \item The behavioral model incorporates both the driver's social cooperativeness and personal driving objectives, which produces rich and interpretable behaviors.
    \item The proposed decision-making module handles the uncertainties in the driving intent using a Bayesian filter and generates smooth and realistic reference trajectories for the downstream low-level vehicle controller.
    \item  Differing from pure learning-based methods, the designed SANN incorporates social-psychological model-based priors. It imitates the behavioral model and is transferable across different traffic conditions while providing better online computation efficiency.
    \item The decision-making module utilizes an interaction-guided decision tree search algorithm, which ensures probabilistic safety and scales linearly with the number of interacting drivers and prediction horizons.
    \item The behavioral model is evaluated in predicting real-world trajectories. This model demonstrates good quantitative accuracy in short-term prediction and provides qualitative long-term behavioral prediction.
    \item The decision-making module is tested in the forced merging scenarios on a comprehensive set of environments without re-tuning the model hyperparameters. The proposed method can safely merge the ego vehicle into the real-world traffic dataset~\cite{highD} faster than the human drivers, and into diverse Carla~\cite{carla} simulated traffic with different traffic conditions.
\end{enumerate}

This paper is organized as follows: In Sec.~\ref{sec:preliminaries}, we describe the model preliminaries, including the vehicle kinematics model, the action space with the lane-change modeling, and the choice of model hyperparameters. In Sec.~\ref{sec:methodModel}, we present the behavioral model, which can be utilized to predict interacting drivers' trajectories given their latent driving intentions. In Sec.~\ref{sec:methodImitation}, we discuss the SANN architecture that imitates the behavioral model and is suitable for online deployment. In Sec.~\ref{sec:methodCtrl}, we introduce the decision-making module of our ego vehicle together with a decision tree search algorithm that incorporates the SANN and improves computation efficiency. In Sec.~\ref{sec:results}, we report the results of real-world trajectory prediction using the behavioral model for the forced merging test on a real-world dataset and in simulations. Finally, conclusions are given in Sec.~\ref{sec:conclusion}.


\section{System and Model Preliminaries}\label{sec:preliminaries}
\begin{figure}[!h]
    \centering
    \includegraphics[width=0.49\textwidth]{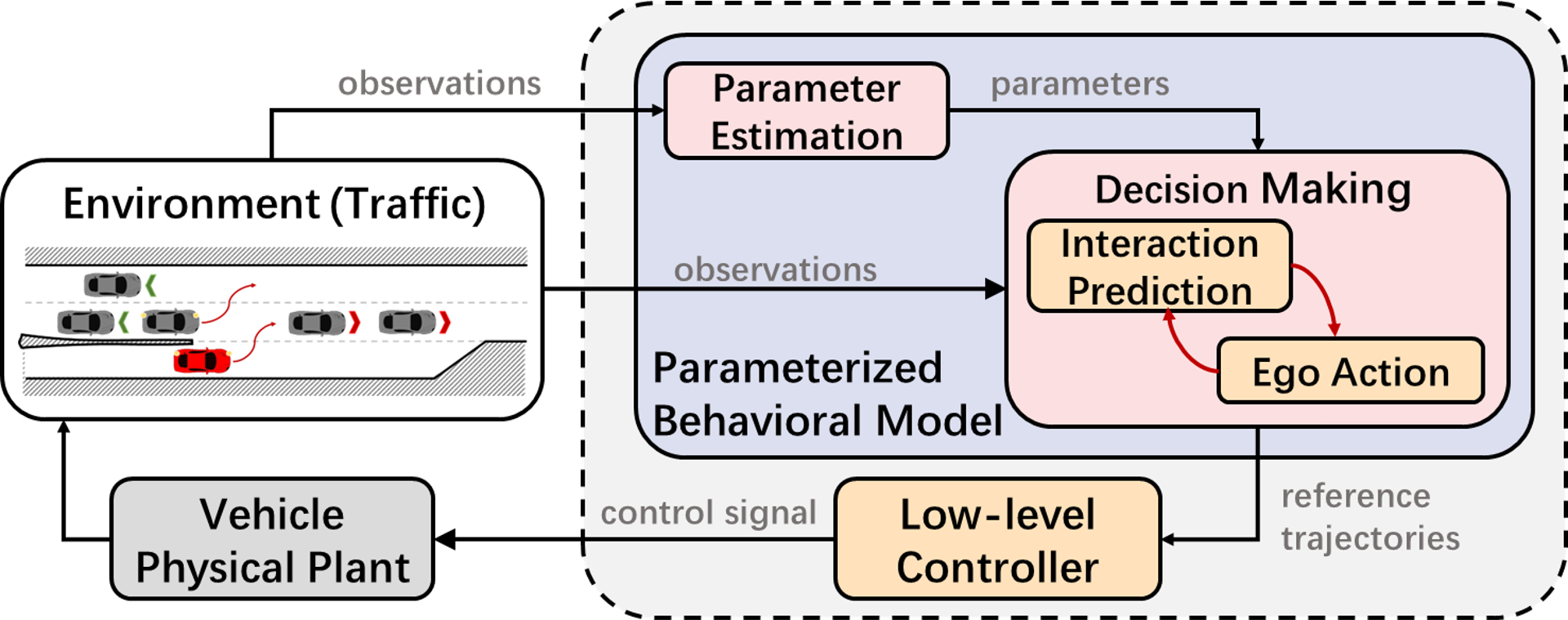}\vspace{-0.5em}
    \caption{Proposed algorithm architecture for autonomous vehicles decision-making and control.}
    \label{fig:systemArch}
\end{figure}

In this paper, we design a modularized algorithm architecture for decision-making and control of the autonomous (ego) vehicle in the forced merging scenario. In this framework (see Fig.~\ref{fig:systemArch}), we develop a parameterized behavioral model for modeling the behavior of interacting drivers. Leveraging this model and observed traffic interactions, we can estimate the latent driving intentions of interacting drivers as model parameters. Thereby, we can predict their future trajectories in response to the action of the ego vehicle. Based on the observations, and the predictions, a high-level decision-making module optimizes a reference trajectory for the ego vehicle merging into the target highway lane while ensuring its safety in the traffic. A low-level controller controls the vehicle throttle and steering angle to track the reference trajectory. 

This chapter first introduces the vehicle kinematics model in Sec.~\ref{subsec:kinematics}. Then, we present a state-dependent action space (in Sec.~\ref{subsec:actionSpace}) of the vehicle, which is a set of trajectories synthesized from the kinematics model. We discuss the detailed lane change trajectory modeling in Sec.~\ref{subsec:lanechange} together with model hyperparameter identification from a naturalistic dataset~\cite{highD}.


\subsection{Vehicle Kinematics}\label{subsec:kinematics}
We use the following continuous-time bicycle model~\cite{rajamani2011vehicle} to represent the vehicle kinematics,
\begin{equation}\label{eq:EoM}
\begin{split}
    \left[\begin{array}{c}
        \dot{x}\\
        \dot{y}\\
        \dot{\varphi}\\
        \dot{v}
    \end{array}\right]
    = \left[\begin{array}{c}
        v\cos(\varphi + \beta)\\
        v\sin(\varphi + \beta)\\
        \frac{v}{l_r}\sin(\beta)\\
        a
    \end{array}\right] + \tilde{w},
    \\
    \beta = \arctan\brk{\frac{l_r}{l_r+l_f}\tan \delta},
\end{split}
\end{equation}
where $x$, $v$, and $a$ are the longitudinal position, velocity, and acceleration of the vehicle center of gravity (CoG), respectively; $y$ is the lateral position of the CoG; $\varphi$ is the heading angle of the vehicle; $\beta$ is the sideslip angle; $\delta$ is the front wheel steering angle; $l_r$ and $l_f$ denote the distance from the vehicle CoG to the front and rear wheel axles, respectively; $\tilde{w}\in\mathbb{R}^4$ is a disturbance representing unmodeled dynamics. 

We assume that all the highway vehicles, together with the ego vehicle, follow this kinematics model. We then derive discrete-time kinematics from~\eqref{eq:EoM} assuming zero-order hold with the sampling period of $\Delta T$ sec. This leads to the discrete-time kinematics model,
\begin{equation}\label{eq:fEoM}
    s_{k+1}^{i} = f\big(s_k^{i}, u_k^{i}\big)  + \tilde{w}_k^{i},\; i = 0, 1, 2,\dots,
\end{equation}
where the subscript $k$ denotes the discrete time instance $t_k=k\Delta T$ sec; the superscript $i$ designates a specific vehicle, where we use $i=0$ to label the ego vehicle and $i=1,2\dots$ for other interacting vehicles; $s_k^{i} = [x^{i}_k,y^{i}_k,\varphi^{i}_k,v^{i}_k]^T$ and $u_k^{i}=[a_k^{i},\delta_k^{i}]^T$ are the state and control vectors of vehicle $i$ at time instance $t_k$. Then, we can use this discrete kinematics model to synthesize vehicle trajectories. Note that there are other vehicle kinematics and dynamics models that could potentially represent vehicle behaviors~\cite{rajamani2011vehicle} more realistically. The model~\eqref{eq:fEoM} was chosen as it provides adequate accuracy for the purpose of decision-making (planning) of the ego vehicle while it is simple and computationally efficient~\cite{kong2015kinematic}.

\subsection{Trajectory Set as Action Space}\label{subsec:actionSpace}
\begin{figure}[!h]
    \centering
    \includegraphics[width=0.49\textwidth]{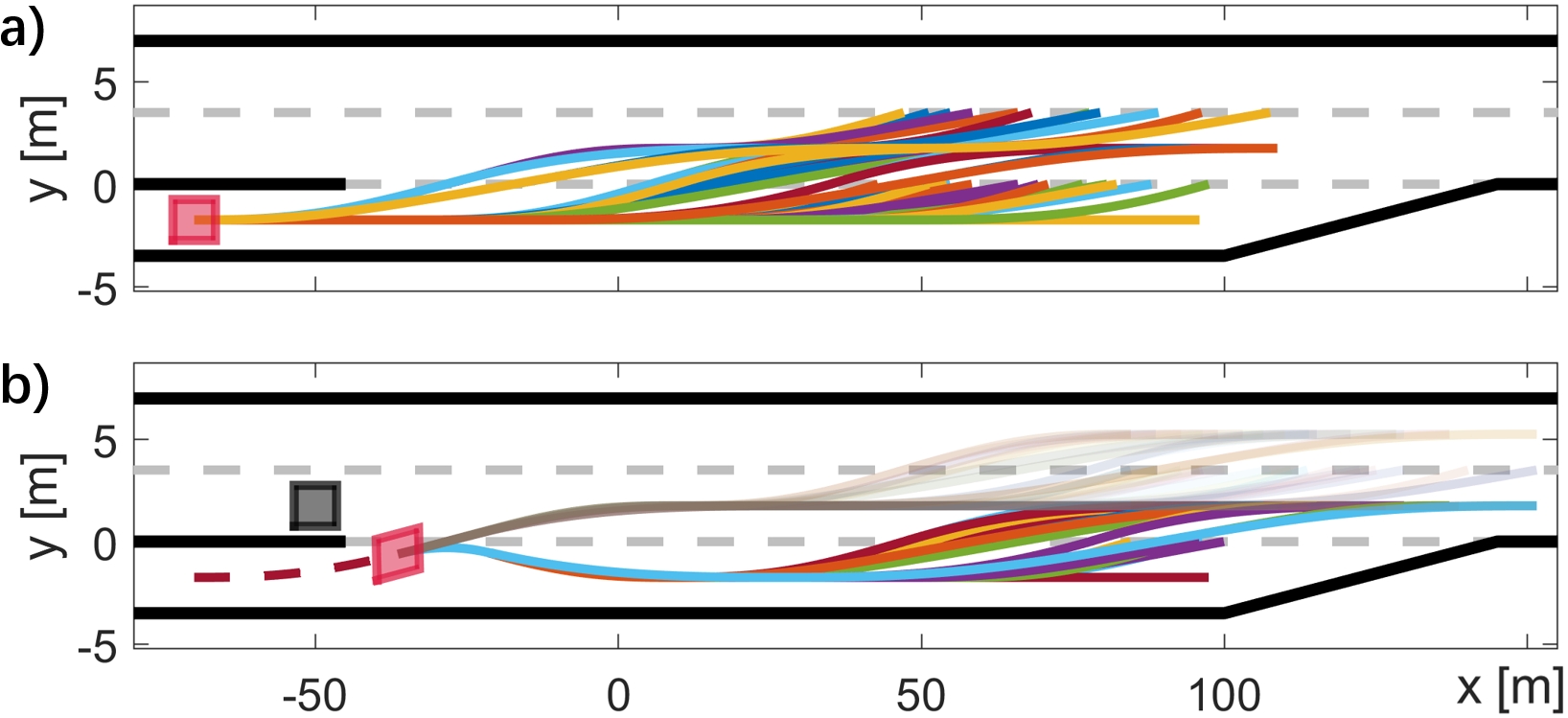}\vspace{-0.5em}
    \caption{Examples of trajectory set $\Gamma(s_k^{i})$ with duration $6\;\rm sec$: (a) A trajectory set of $M=109$ encompasses behaviors of lane keeping, lane change, and coupled longitudinal and lateral behavior (e.g., lane change with longitudinal acceleration/deceleration). (b) A trajectory set of $M=129$ contains actions of lane change abortion and re-merge after aborting the previous lane change. Other normal lane change trajectories are in semi-transparent lines. Likewise, the lane change abortion behaviors are also coupled with various longitudinal acceleration/deceleration profiles.}
    \label{fig:trajectorySets}
\end{figure}

Given the initial vehicle state and the kinematics model~\eqref{eq:fEoM} neglecting the disturbance $\tilde{w}_k^i$ and using different control signal profiles, we can synthesize various trajectories with duration $N\Delta T$ sec for trajectory prediction and planning. For the $i$th vehicle at time $t_k$, we assume that vehicle's action space is $\Gamma(s_k^{i})=\set{\gamma_{m}(s_k^{i})}_{m=1}^{M}$ where each individual element $\gamma^{(m)}(s_k^{i}) = \set{s_{n}^{i}}_{n=k}^{k+N+1}$ is a trajectory of time duration $N\Delta T$ sec synthesized using a distinct control sequence $\set{u_{n}^i}_{n=k,\dots,k+N}$ via the kinematics model~\eqref{eq:fEoM}. We select $225$ different control sequences such that the number of considered trajectories is finite, i.e., $M\leq 225$ for all $\Gamma(s_k^{i})$ and all initial state $s_k^{i}$. As shown in Fig.~\ref{fig:trajectorySets}, the action space $\Gamma(s_k^{i})$ depends on the current vehicle state $s_k^{i}$ for two reasons: With a fixed control sequence $\set{u_{n}^i}_{n}$, the resulted trajectory from~\eqref{eq:fEoM} varies with the initial condition $s_k^{i}$; A safety filter is implemented for $\Gamma(s_k^{i})$ such that all trajectories intersect with the road boundaries are removed, which is also dependent on $s_k^i$. The chosen $225$ control sequences generate trajectories that encompass a set of plausible driving behaviors (see Fig.~\ref{fig:trajectorySets}) and suffice tasks of trajectory prediction and planning. Note that the trajectory set can be easily enlarged with more diverse control sequences if necessary.

Meanwhile, we assume a complete lane change takes $T_{\text{lane}}=N_{\text{lane}}\Delta T$ sec to move $w_{\text{lane}}$ meters from the center line of the current lane to that of adjacent lanes. We set $N_{\text{lane}}<N$ to allow trajectory sets to contain complete lane change trajectories. As shown in Fig.~\ref{fig:trajectorySets}a), the trajectory set comprises 109 regular driving trajectories for an on-ramp vehicle that is intended to merge. This trajectory set considers varieties of driver's actions such as lane keeping with longitudinal accelerations/decelerations, merging with constant longitudinal speed, merging with longitudinal acceleration/deceleration, accelerating/decelerating before or after merging, etc. Moreover, we also include the behavior of aborting lane change (see Fig.~\ref{fig:trajectorySets}b). For a lane-changing vehicle, this is a regular ``change-of-mind" behavior to avoid collision with nearby highway vehicles. For longitudinal behaviors, we also assume speed and acceleration/deceleration limits of $[v_{\min},v_{\max}]$ and $[a_{\min},a_{\max}]$ for all vehicle at all times. Namely, the trajectory sets and control sequences satisfy $a_{n}^i\in[a_{\min},a_{\max}]$ and $v_{n}^{i}\in[v_{\min},v_{\max}]$ for all $s_{n}^{i}\in \gamma^{(m)}(s_k^{i}), n=k,\dots,k+N+1$ and for all 
$\gamma^{(m)}(s_k^{i})\in\Gamma(s_k^{i}), m=1,\dots,M(s_k^{i})$. The speed limits are commonly known quantities on highways and the longitudinal acceleration/deceleration is typically limited by the vehicle's performance limits.

\subsection{Trajectory Hyperparameters and Lane Change Behavior}\label{subsec:lanechange}
\begin{figure}[ht!]
    \centering
    \includegraphics[width=0.49\textwidth]{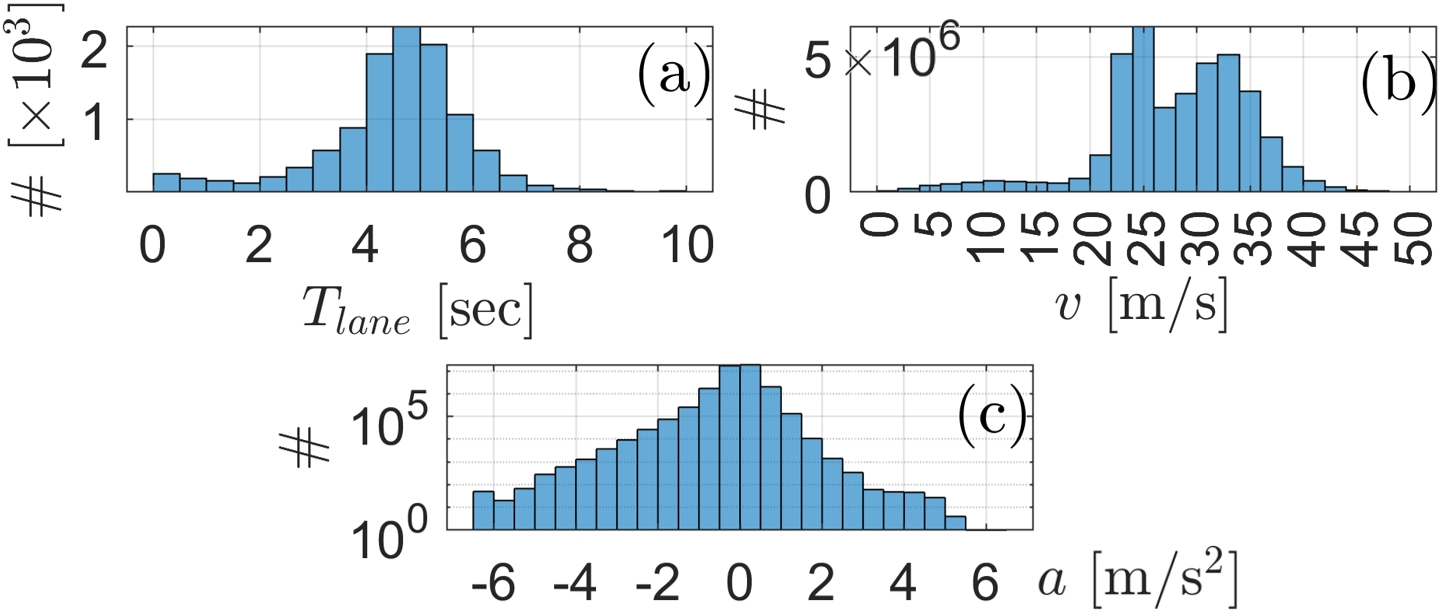}\vspace{-0.5em}
    \caption{Histogram of vehicle driving statistics in the High-D dataset \cite{highD}: (a) Time duration for a complete lane change. (b) Longitudinal velocity. (c) Longitudinal acceleration/deceleration (y-axis in log scale).}
    \label{fig:param_histogram}
\end{figure}

We use a naturalistic highway driving High-D~\cite{highD} dataset to identify the model hyperparameters, i.e., $v_{\min}$, $v_{\max}$, $a_{\min}$, $a_{\max}$, $w_{\text{lane}}$, and $T_{\text{lane}}$. The statistics visualized in Fig.~\ref{fig:param_histogram} are obtained from data of 110,500 vehicles driven over 44,500 kilometers. The minimum speed is set to $v_{\min}=2\;\rm m/s$ since most of the vehicles have speeds higher than that and the maximum speed limit of the dataset is $v_{\max}=34\;\rm m/s$. The majority of longitudinal accelerations and decelerations of High-D vehicles are within the range of $[a_{\min},a_{\max}]=[-6, 6]\;\rm m/s^2$. The lane width $w_{\text{lane}}=3.5\;\rm m$ as in High-D dataset. We select $T_{\text{lane}}=N_{\text{lane}}\Delta T=4\;\rm sec$ since most of the High-D vehicles take between $4$ and $6\;\rm sec$ to change lanes. We keep these hyperparameters fixed for the following discussion and experiments. Note that these parameters can be identified similarly to different values in other scenarios if necessary.

In terms of lane change behaviors, given an acceleration sequence $\set{a_k^i}_{k}$, we can derive the steering profile $\set{\delta_k^i}_{k}$ of a lane change trajectory from 5th order polynomials~\cite{TomiPolyLaneChange},
\begin{equation}\label{eq:laneChangePoly}
\begin{split}
    x(t|\set{p_j}) = p_0 + p_1 t + p_2 t^2 + p_3 t^3 + p_4 t^4 + p_5 t^5,\\
    y(t|\set{q_j}) = q_0 + q_1 t + q_2 t^2 + q_3 t^3 + q_4 t^4 + q_5 t^5,
\end{split}
\end{equation}
which represents a vehicle lane change between time $t=0$ and $t=T_{\text{lane}}\;\rm sec$. Such lane change trajectories are commonly used in vehicle trajectory planning and control~\cite{yue2018robust,you2015trajectory}. 

\begin{figure}[ht!]
    \centering
    \includegraphics[width=0.49\textwidth]{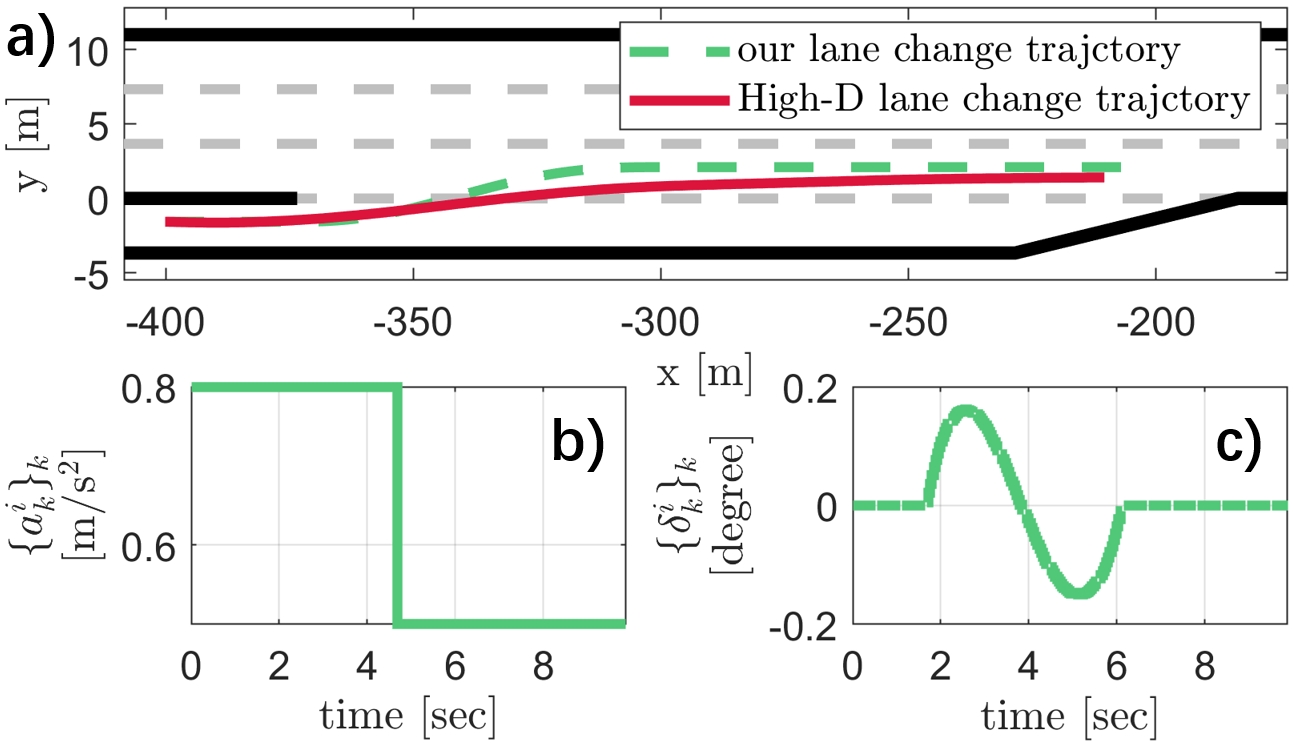}\vspace{-0.5em}
    \caption{A lane change trajectory synthesized using a given acceleration sequence: (a) Synthesized trajectory using our algorithm compared with a real-world lane change trajectory in the High-D dataset. (b) Designed acceleration sequence $\set{a_k^i}_{k}$. (c) Derived steering sequence $\set{\delta_k^i}_{k}$ from~\eqref{eq:fEoM}.}
    \label{fig:laneChangeComp}
\end{figure}

Suppose, without loss of generality, that the lane change starts and ends at $t_0=0$ and $t_{N_{\text{lane}}}=T_{\text{lane}}\;\rm sec$, respectively. The following procedure is utilized to determine the trajectory and steering profile during a lane change: At time $t_k=k\Delta T\;\rm sec$, given vehicle state $s_{k}^i= [x^{i}_k,y^{i}_k,\varphi^{i}_k,v^{i}_k]^T$ and lateral target lane center $y_{\text{target}}$, we first solve for the coefficients $\set{p_{k,j}}$ and $\set{q_{k,j}}$ in~\eqref{eq:laneChangePoly} from the following two sets of boundary conditions,
\begin{equation}\label{eq:polyBoundaryCondition}
\begin{array}{c}
\left\{
\begin{array}{ccc}
    x(t_k) = x^i_k, & \dot{x}(t_k) = v^i_k, &\ddot{x}(t_k) = a^i_k,\\
    y(t_k) = y^i_k, & \dot{y}(t_k) = \dot{y}^i_k, &\ddot{y}(t_k) = \ddot{y}^i_k,
\end{array}
\right.
\\\\
\left\{
\begin{array}{c}
    x(T_{\text{lane}}) = x^i_k+v^i_k(T_{\text{lane}}-t_k) \\+\frac{1}{2}a^i_k(T_{\text{lane}}-t_k)^2,\\
    \dot{x}(T_{\text{lane}}) = v^i_k+a^i_k(T_{\text{lane}}-t_k),\;
    \ddot{x}(T_{\text{lane}}) = a^i_k,\\ 
    y(T_{\text{lane}}) = y_{\text{target}},\;\dot{y}(T_{\text{lane}}) = 0,\; \ddot{y}(T_{\text{lane}}) =0, \\
\end{array}
\right.
\end{array}
\end{equation}
where we assume initial/terminal conditions $\dot{y}(0) = \dot{y}(T_{\text{lane}}) = 0$ and $\ddot{y}(0) = \ddot{y}(T_{\text{lane}}) = 0$, i.e., zero lateral velocity and acceleration at the beginning and the end of a lane change. Recursively, initial conditions $\dot{y}(t_k) = \dot{y}^i_k$ and $\ddot{y}(t_k) = \ddot{y}^i_k$ at step $k$ can be computed using \eqref{eq:laneChangePoly} with the coefficients $\set{q_{k-1,j}}$ at previous step $k-1$; and we assume a constant longitudinal acceleration $a^i_k$ throughout the lane change process. Then, we can compute $s_{k+1}^i=[x_{k+1}^i, y_{k+1}^i, \varphi_{k+1}^i, v_{k+1}^i]^T$ from~\eqref{eq:laneChangePoly} using the following equations,
\begin{equation}
\begin{array}{c}
    x_{k+1}^i = x(t_{k+1}|\set{q_{k,j}}_j), y_{k+1}^i = y(t_{k+1}|\set{q_{k,j}}_j),\\
    \varphi_{k+1}^i = \arctan\Big(\dot{y}(t_{k+1}|\set{q_{k,j}}_j)/\dot{x}(t_{k+1}|\set{q_{k,j}}_j)\Big),\\
    v_{k+1}^i =\dot{x}(t_{k+1}|\set{q_{k,j}}_j).
\end{array}    
\end{equation}
Repeating this procedure for $k=0,1,\dots,N_{\text{lane}}-1$, we can synthesize a smooth lane change trajectory $\set{s_k^i}_{k=0,\dots,N_{\text{lane}}}$ with corresponding acceleration sequences $\set{a_k^i}_{k=0,\dots,N_{\text{lane}}-1}$. Fig.~\ref{fig:laneChangeComp} illustrates this approach to the lane change modeling. Given an acceleration sequence $\set{a_k^i}_{k}$ (see Fig.~\ref{fig:laneChangeComp}b) from one of the $225$ control sequences $\set{u_k^i}_{k}$, we leverage~\eqref{eq:laneChangePoly} and produce a smooth lane change trajectory that qualitatively matches with a real-world (High-D) lane change trajectory. Meanwhile, the resulting steering angle profile $\set{\delta_k^i}_{k}$ (see Fig.~\ref{fig:laneChangeComp}c) is similar to those from human driving~\cite{salvucci2002time, salvucci2006modeling}.

\section{Social Behavior Modeling}\label{sec:methodModel}
In this section, we model the two components of drivers' driving incentives that motivate them to take action from the trajectory sets defined in Sec.~\ref{sec:preliminaries}. The first component consists of each individual driver's objectives as a personal reward in Sec.~\ref{subsec:objectives}. The second component uses an SVO-based reward to incorporate the drivers' social cooperativeness (see Sec.~\ref{subsec:reward}). In Sec.~\ref{subsec:behaviorModel}, we integrate this reward model into the interacting vehicle’s decision-making process. 

\subsection{Driver's Driving Objectives and Personal Rewards}\label{subsec:objectives}
Similar to the previous work~\cite{ACC2024}, we model the personal reward of the $i$th driver who interacts with an adjacent vehicle $j$ using the following formula, 
\begin{equation}\label{eq:reward_r}
\begin{array}{l}
r\Big(\gamma_{n_1}^{n_2}(s_k^{i}), \gamma_{n_1}^{n_2}(s_k^{j})| w^i\Big)  = 
\neg c\Big(\gamma_{n_1}^{n_2}(s_k^{i}),\gamma_{n_1}^{n_2}(s_k^{j})\Big)\cdot \\
\left[
\begin{array}{ccc}
    h(s^i_{k+n_2}, s^j_{k+n_2}) &
    \tau(s^i_{k+n_2}) &
    e\brk{\gamma_{n_1}^{n_2}(s_k^{i})}
\end{array}
\right]\cdot w^i
\end{array},
\end{equation}
where $s^i_k, s^j_k$ are the current states of the vehicles $i,j$; $\gamma_{n_1}^{n_2}(s_k^{i})=\{s_{k+n}^{i}\}_{n=n_1}^{n_2}\subset\gamma(s_k^{i})$ is a segment of the trajectory $\gamma(s_k^{i})\in\Gamma(s_k^i)$, and $0\leq n_1\leq n_2\leq N+1$; $\gamma_{n_1}^{n_2}(s_k^{j})$ is defined likewise; $\neg$ is the logical negative operator; $w^i\in\mathbb{R}^3$ is a vector of weights so that the personal reward is a weighted summation of personal objectives $h$, $\tau$, and $e$. Here, $c$, $h$, $\tau$, and $e$ are four functions that capture different aspects of drivers' driving objectives:
\begin{enumerate}
    \item Collision avoidance $c\in\set{0,1}$: $c\big(\gamma_{n_1}^{n_2}(s_k^{i}),\gamma_{n_1}^{n_2}(s_k^{j}) \big)=1$ implies vehicle $i$ following trajectory $\gamma_{n_1}^{n_2}(s_k^{i})$ collides with vehicle $j$ which follows trajectory $\gamma_{n_1}^{n_2}(s_k^{j})$, and $c=0$ indicates that two vehicles' trajectories are free of collision with each other. This is used to penalize collisions between trajectories.
    \item Safety consciousness $h\in[0,1]$: If vehicle $j$ is the leading vehicle of vehicle $i$, $h(s^i_{k+n_2}, s^j_{k+n_2})$ computes a normalized Time-to-Collision $(TTC)$ at the end of their corresponding trajectories $\gamma_{n_1}^{n_2}(s_k^{i}), \gamma_{n_1}^{n_2}(s_k^{j})$. A larger $h$ implies a larger TTC with the leading vehicle. The safety consciousness can encourage vehicles to keep an appropriate headway distance and be conscious of potential collisions.
    \item Travelling time $\tau\in[0,1]$: $\tau(s^i_{k+n_2})$ measures the closeness between the vehicle's final state in the trajectory $\gamma_{n_1}^{n_2}(s_k^{i})$ with its destination, where a larger value implies shorter distance to the goal. Including $\tau$ in the reward reflects the objective of shortening the traveling time, e.g., merging to the highway as soon as possible for the on-ramp vehicles.
    \item Control effort $e\in[0,1]$: $e\brk{\gamma_{n_1}^{n_2}(s_k^{i})}$ takes a lower value if $\gamma_{n_1}^{n_2}(s_k^{i})$ is a lane-changing trajectory segment or generated with longitudinal acceleration/deceleration. The control effort captures drivers' desire to keep the lane and constant speed to avoid both longitudinal and lateral maneuvers.
\end{enumerate}
We refer the readers to our previous work~\cite{ACC2024} for more detailed descriptions of the four functions. Similar methods that model the driver's driving objectives have also been reported in~\cite{brechtel2011probabilistic, wei2011point, chen2019model, liu2022interaction, li2017game}. The weight $w^i$ is the latent model parameter in the reward function $r(\cdot|w^i)$. Different weights reflect distinct personal goals and, therefore, embed various driving behaviors. For example, a driver considering personal reward with $w^i=[0,0,1]^T$ may keep the lane and drive at a constant speed, thereby maximizing the reward via minimizing the control effort. Another driver with weights $w^i=[1,0,0]^T$ tries to maximize the headway distance and might change lanes to overtake a leading vehicle if there is one. 

\subsection{Social Value Orientation and Multi-modal Reward}\label{subsec:reward}
The personal reward function $r(\cdot|w^i)$ captures drivers' decision-making as maximizing their own gain in the traffic interaction. However, this model does not encode the behaviors of cooperation and competition. For example, a highway driver observing the merging intention of an on-ramp vehicle might slow down to yield. In social psychology studies~\cite{messick1968motivational, liebrand1984effect}, the notion of Social Value Orientation (SVO) was proposed to model this cooperative/competitive behavior in experimental games, and it has more recently been applied to autonomous driving~\cite{DanielaRusSVO}. Taking inspiration from this work, we use the driver's SVO to incorporate the personal reward with the driver's tendency toward social cooperation.

To this end, we assume each vehicle $i$ interacts with the adjacent vehicle $j\in A(i)$, where $A(i)$ contains indices of all the adjacent vehicles around vehicle $i$. We model driver $i$'s intention using a multi-modal reward function of the form,
\begin{equation}\label{eq:reward_R}
\begin{array}{l}
    R \Big(\gamma_{n_1}^{n_2}(s_k^{i}), \boldsymbol{\gamma}_{n_1}^{n_2}(\boldsymbol{s}_k^{-i})| \sigma^i, w^i \Big) \\
    = \alpha(\sigma_i) \cdot  r\Big(\gamma_{n_1}^{n_2}(s_k^{i}), \gamma_{n_1}^{n_2}(s_k^{j})| w^i\Big)\\
    +\beta(\sigma_i)\cdot \E_{j\in A(i)} \bsq{ r\Big(\gamma_{n_1}^{n_2}(s_k^{j}), \gamma_{n_1}^{n_2}(s_k^{i})| w^j \Big)},
\end{array}
\end{equation}
where $\boldsymbol{s}_k^{-i}=[s_k^0, s_k^1, s_k^2, \dots]$ is the aggregated state of all the adjacent vehicles of vehicle $i$; $\boldsymbol{\gamma}_{n_1}^{n_2}(\boldsymbol{s}_k^{-i})=[\gamma_{n_1}^{n_2}(s_k^0),\gamma_{n_1}^{n_2}(s_k^1),\gamma_{n_1}^{n_2}(s_k^2),\dots]$ concatenates one possible trajectory segment $\gamma_{n_1}^{n_2}(s_k^j)$ for each vehicle $j\in A(i)$; The SVO $\sigma^i$ is another latent model parameter. It takes one of the four values corresponding to four SVO categories and the values of $\alpha(\sigma^i)$ and $\beta(\sigma^i)$ are specified as follows
\begin{equation}\label{eq:sigma}
    (\alpha, \beta) =\left\{
    \begin{array}{cl}
         (0,1) &  \text{if } \sigma^i = \text{ ``altruistic'',}\\
         (1/2, 1/2) &  \text{if } \sigma^i = \text{ ``prosocial'',}\\
         (1,0) &  \text{if } \sigma^i = \text{ ``egoistic'',}\\
         (1/2, -1/2) &  \text{if } \sigma^i = \text{ ``competitive''.}\\ 
    \end{array}
    \right.
\end{equation}
In \eqref{eq:reward_R}, $\alpha(\sigma^i)$ weighs the self-reward while $\beta(\sigma^i)$ weighs an averaged reward to the other vehicles. We also note that the weight $w^j$ is an internal parameter of vehicle $j$ and is a latent variable affecting the decision of vehicle $i$. Similar to~\cite{ACC2024}, we assume $w^j=[1/3, 1/3, 1/3]$ in Eq.~\eqref{eq:reward_R} for $j \in A(i)$. The rationale behind this assumption is that an altruistic or prosocial (or competitive) driver of vehicle $i$ is likely to cooperate (or compete) with other drivers in all three objectives if they do not know others' actual intentions.  

Using this multi-modal reward, we can model each driver's intention to achieve their personal objectives (reflected in $w^i$) and, to a certain extent, cooperate with others (encoded in $\sigma^i$). For example, suppose two highway drivers with the same personal weights $w^i=[0,0,1]^T$ encounter a merging on-ramp vehicle. A ``egoistic'' highway driver values the control effort heavily and, therefore is likely to keep the lane at a constant speed and ignore the merging vehicle. On the contrary, a ``prosocial" highway driver might consider changing lanes or slowing down to promote on-ramp merging action such that the net reward in~\eqref{eq:reward_R} is larger.

\subsection{Driving Behavior Model}\label{subsec:behaviorModel}
Using the multi-modal reward, we can decode/infer drivers' intentions from their actions/trajectories, which can be represented by the model parameters $w^i,\sigma^i$. We formalize this process into a behavioral model,
\begin{equation}\label{eq:behavior_model_MPC}
\begin{array}{c}
    \gamma^*(s_k^i) = \argmax\limits_{\gamma(s_k^i)\in \Gamma(s_k^i)}\; Q\brk{\boldsymbol{s}_k^{-i}, \gamma(s_k^i)| \sigma^i, w^i},
\end{array}
\end{equation}
where $\gamma^*(s_k^i)$ is the resulting reference trajectory for vehicle $i$ and $Q$ denotes the corresponding cumulative reward function. This cumulative reward admits the following form,
\begin{align}\label{eq:behavior_model_Q}
\begin{aligned}
     & Q\brk{\boldsymbol{s}_k^{-i}, \gamma(s_k^i)| \sigma^i, w^i} = \E\limits_{\boldsymbol{\gamma}(\boldsymbol{s}_k^{-i})\in\boldsymbol{\Gamma}(\boldsymbol{s}_k^{-i})}\\
    &\bsq{\sum\limits_{n=0}^{{\scriptscriptstyle  \lfloor N/N'\rfloor}} \lambda^{n} R \Big(\gamma{\scriptscriptstyle  _{nN'}^{(n+1)N'}}(s_k^{i}), \boldsymbol{\gamma}{\scriptscriptstyle  _{nN'}^{(n+1)N'}}(\boldsymbol{s}_k^{-i})| \sigma^i, w^i \Big) },
\end{aligned}
\end{align}
where $\lambda\in(0,1)$ is a discount factor; the summation is a cumulative reward of vehicle $i$ over a $N\Delta T$ sec look-ahead/prediction horizon, and this reward is obtained according to \eqref{eq:reward_R}; $N'\Delta T$ ($N'<N$) in second denotes the sampling period that the driver updates its decision; $\lfloor x\rfloor$ denotes the largest integer lower bound of $x\in \R$; the expectation averages the reward over all possible aggregated trajectories in the set,
\begin{equation}
    \boldsymbol{\Gamma}(\boldsymbol{s}_k^{-i})=\set{\boldsymbol{\gamma}(\boldsymbol{s}_k^{-i}):\gamma(s_k^j)\in\Gamma(s_k^j),j\in A(i)}.
\end{equation}

After obtaining the optimal $\gamma^*(s_k^i)$ using \eqref{eq:behavior_model_MPC}, we compute the control signal $u_n^{i}=[a_n^{i},\delta_n^{i}]^T$ at each time step $n=k,\dots,k+N'$ to track this reference trajectory $\gamma^*(s_k^i)$ for one sampling period $N'\Delta T$ sec. Then, we update the reference trajectory using \eqref{eq:behavior_model_MPC} after $N'\Delta T$ sec. Eventually, using the behavioral model \eqref{eq:behavior_model_MPC}, we formulate the decision-making process of a highway driver motivated by the reward~\eqref{eq:behavior_model_Q} and a combination of social psychological model parameters $\sigma^i, w^i$, while the driving behaviors are formalized as a receding-horizon optimization-based trajectory-tracking controller with a horizon of $\lfloor N/N'\rfloor$. However, solving this problem online can be computationally demanding. We can use a neural network to learn the solutions of~\eqref{eq:behavior_model_MPC} offline from a dataset, thereby imitating this behavioral model for online deployment.

\section{Interaction-Aware Imitation Learning with Attention Mechanism}\label{sec:methodImitation}
Based on~\eqref{eq:behavior_model_MPC}, the decision-making process of vehicle $i$ is deterministic given $s^i_k$, $\boldsymbol{s}_k^{-i}$. Instead of learning a one-hot encoding and to include stochasticity in the decision-making process, we prescribe a policy distribution from~\eqref{eq:behavior_model_Q} using a softmax decision rule \cite{sutton2018reinforcement} according to, 
\begin{equation}\label{eq:behavior_model_policy}
\begin{array}{l}
    \pi \brk{\gamma(s_k^i) | \sigma^i, w^i, s_k^i, \boldsymbol{s}_k^{-i}} \\
    \propto \exp \bsq{ Q\brk{\boldsymbol{s}_k^{-i}, \gamma(s_k^i)| \sigma^i, w^i} }.
\end{array}
\end{equation}
Note that $\pi$ takes values in $\R^{225}$, where we assign zero probabilities in $\pi$ for the unsafe trajectories $\gamma_{\text{unsafe}}(s_k^{i})\notin\Gamma(s_k^{i})$ filtered out in Sec.~\ref{subsec:actionSpace}.
Then, we can learn a neural network mapping $\pi_{NN}$ for imitating the actual behavioral model $\pi$ using minimization of a modified Kullback–Leibler divergence according to the loss function,
\begin{equation}\label{eq:kl_loss}
\begin{array}{l}
    \mathcal{L}\brk{\pi, \pi_{NN}| \sigma^i, w^i, s_k^i, \boldsymbol{s}_k^{-i}}
    \\
    = \sum\limits_{m=1}^{225} 
    \Big\{ 
    \pi \brk{\gamma^{(m)}(s_k^i)}
    \cdot
    \log\Big[\pi \brk{\gamma^{(m)}(s_k^i)} + \epsilon\Big]
    \\
    - 
    \pi \brk{\gamma^{(m)}(s_k^i)}
    \cdot
    \log\bsq{\pi_{NN} \brk{\gamma^{(m)}(s_k^i)} + \epsilon}
    \Big\},
\end{array}
\end{equation}
where a positive constant $\epsilon\ll 1$ is chosen to avoid zero probability inside the logarithm for numerical stability, and for simplicity, we omit the terms $\sigma^i, w^i, s_k^i, \boldsymbol{s}_k^{-i}$ in the notation of $\pi$, $\pi_{NN}$. This loss function $\mathcal{L}(\cdot)\geq 0 $ measures the similarity between two discrete probability distributions where smaller loss implies more similar distributions. 

\begin{figure}[ht!]
    \centering
    \includegraphics[width=0.45\textwidth]{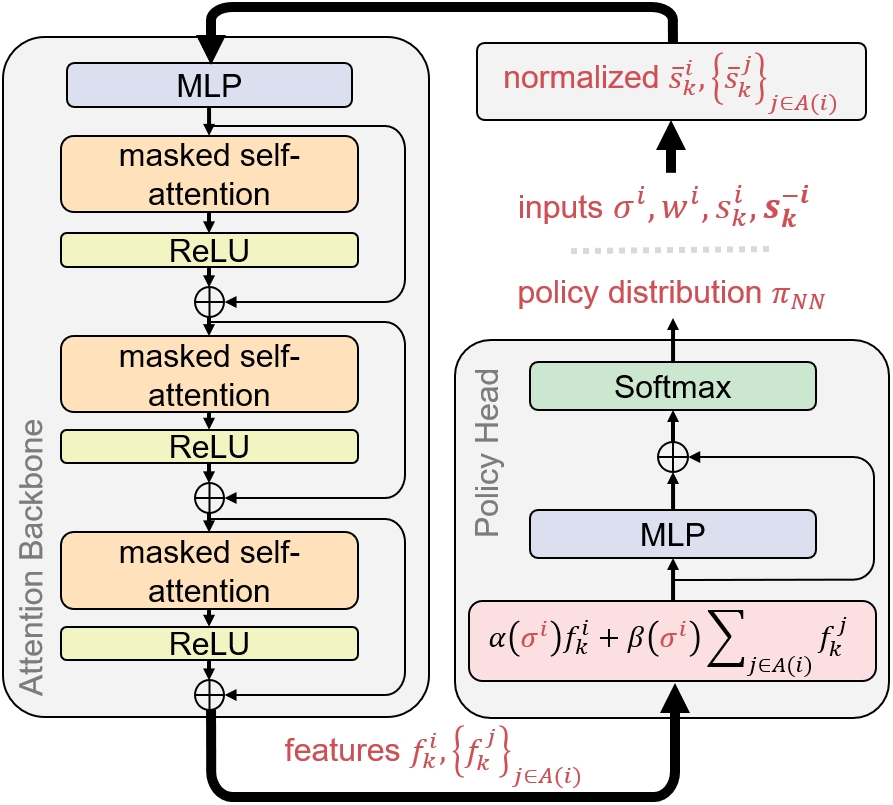}\vspace{-0.5em}
    \caption{Schematic diagram of our SANN architecture: The attention backbone takes the normalized input vectors and produces their corresponding feature vectors via the attention mechanism~\cite{transformer}. The policy head fits a policy distribution $\pi_{NN}$ from the resulting feature vectors incorporating the driver $i$'s social value orientation $\sigma^i$.}
    \label{fig:net_arch}
\end{figure}

We design a Social-Attention Neural Network (SANN) architecture (see Fig.~\ref{fig:net_arch}) that comprises three components: The input normalization (Sec.~\ref{subsec:net_normal}) derives a set of normalized vectors from inputs $\sigma^i, w^i, s_k^i, \boldsymbol{s}_k^{-i}$ using the highway structural information. Then, we generate a set of feature vectors via an interaction-aware learning process using the attention backbone in Sec.~\ref{subsec:net_backbone}, and we present the attention mechanism in Sec.~\ref{subsec:net_att}. Finally, using the learned feature vectors, the policy head imitates the policy distribution $\pi$ from the behavioral model (Sec.~\ref{subsec:net_head}). 

\subsection{Input Normalization}\label{subsec:net_normal}
Given different lane dimensions as labeled in Fig.~\ref{fig:forcedMergingProblem}, we aim to normalize the inputs $\sigma^i, w^i, s_k^i, \boldsymbol{s}_k^{-i}$ accordingly to facilitate the neural network training. The normalization produces input vectors $\bar{s}_k^i$ for vehicle $i$ and a set of vectors $\{\bar{s}_k^j\}$ for $j\in A(i)$ according to,
\begin{equation}\label{eq:NN_normal}
    \bar{s}_k^{\iota} = 
    \bsq{
    \begin{array}{c}
        \brk{x_k^{\iota} - l^{\iota}/2 - x_{\text{ramp}}}/l_{\text{ramp}}\\
        \brk{x_k^{\iota} + l^{\iota}/2 - x_{\text{ramp}}}/l_{\text{ramp}}\\
        \brk{y_k^{\iota} - y_{\text{ramp}}}/w_{\text{ramp}}\\
        (v_k^{\iota}-v_{\min})/(v_{\max}-v_{\min})\\
        w^{\iota}
    \end{array}
    }, \iota \in \set{i}\cup A(i),
\end{equation}
where $l^{\iota}$ is the wheelbase length of vehicle $\iota$; the first two elements of the feature vector~\eqref{eq:NN_normal} are the normalized longitudinal coordinates of the vehicle rear end and front end; $w^{\iota}=[1/3, 1/3, 1/3]$ for all $\iota\in A(i)$ per Sec.~\ref{subsec:reward}.
\subsection{Attention Backbone}\label{subsec:net_backbone}
We first use a multi-layer perceptron (MLP), i.e., a fully connected neural network, to expand the dimension of the inputs $\{\bar{s}_k^{\iota}\}$ individually from $\R^7$ to $\R^{225}$ according to,
\begin{equation}\label{eq:nn_mlp}
    z_{\ell}=\sigma_{\text{ReLU}}\left(W_{\ell}z_{\ell-1}+b_{\ell}\right),~~ \ell=1,\dots,L,
\end{equation}
where $W_{\ell}$ and $b_{\ell}$ are the network parameters of the $\ell$th layer; $\sigma_{\text{ReLU}}(z)=\max\set{0,z}$ is an element-wise ReLU activation function; $L\in\Z$ is the number of layers; the inputs are $z_0=\bar{s}_k^{\iota}\in\R^5$, $\iota \in \set{i}\cup A(i)$; the outputs of the MLP are learned vectors $z^{\iota}_k=z_L\in\R^{225}$, $\iota \in \set{i}\cup A(i)$. Then, we combine the learned vectors $\{z^{\iota}_k\}_{\iota}$, into one matrix $Z=[z^{i}_k, \dots, z^{j}_k, \dots]^T$ where each row corresponds to a learned feature vector $(z^{\iota}_k)^T$. The row dimension of $Z$ can vary with the numbers of interacting vehicles in $A(i)$, which is undesirable for forming a batch tensor in neural network training~\cite{pytorch}. Thus, we consider a maximum of $N_z-1$ adjacent vehicles in $A(i)$ such that we can append zero rows to $Z$ and construct $Z\in\R^{N_z\times 225}$. Moreover, we use a mask matrix $H\in\R^{N_z\times N_z}$ to mark down the indices of the appended rows for the latter masked self-attention process. The element of $H$ in the $i$th row and $j$th column attains $H_{i,j}=-\infty$ if the $i$th or $j$th row vector in $Z$ is an appended zero vector, and obtains $H_{i,j}=0$ otherwise.

Subsequently, we pass $Z$ through three identical cascaded blocks (see Fig.~\ref{fig:net_arch}) using the following formula,
\begin{equation}\label{eq:nn_backbone}
\begin{array}{c}
    \bar{Z}_{\ell} = {\tt Attention} (Z_{\ell-1}| W_{Q,\ell}, W_{K,\ell}, W_{V,\ell}),\\
    Z_{\ell}=\sigma_{\text{ReLU}}(\bar{Z}_{\ell})+ Z_{\ell-1},~~\ell=1,2,3,
\end{array}
\end{equation}
where ${\tt Attention}(\cdot)$ denotes the masked self-attention block~\cite{transformer}; $W_{Q,\ell}\in\R^{225\times|Q|}$, $W_{K,\ell}\in\R^{225\times|Q|}$, and $W_{V,\ell}\in\R^{225\times|V|}$ are the parameters named query, key, and value matrix of the $\ell$th masked self-attention; the inputs are $Z_0=Z$ and each layer $\ell=1,2,3$ produces $Z_{\ell}\in \R^{N_z\times 225}$; the summation outside $\sigma_{\text{ReLU}}$ corresponds to the bypass connection in Fig.~\ref{fig:net_arch} from the beginning of a masked self-attention block to the summation symbol $\oplus$. This bypass connection is called residual connection~\cite{resnet} and is designed to mitigate the vanishing gradient issue in the back-propagation of deep neural networks. We chose to cascade three such blocks via trading-off between empirical prediction performance and computational cost in comparison with those using different numbers of this block.

\subsection{Attention Mechanism}\label{subsec:net_att}
In the $\ell$th attention block~\eqref{eq:nn_backbone}, we leverage the attention mechanism to interchange information between row vectors of the matrix $Z_{\ell-1}$ in the learning process. Specifically, the $\ell$th masked self-attention block can be represented using the following set of equations,

\begin{subequations}\label{eq:nn_att}
\noindent\centering
\begin{minipage}{0.23\textwidth}
\begin{equation}\label{eq:nn_att_Q}
    Q_{\ell} = Z_{\ell-1}W_{Q,\ell},
\end{equation}
\end{minipage}
\hfill
\begin{minipage}{0.23\textwidth}
\begin{equation}\label{eq:nn_att_K}
    K_{\ell} = Z_{\ell-1}W_{K,\ell},
\end{equation}
\end{minipage}
\begin{equation}\label{eq:nn_att_V}
    V_{\ell} = Z_{\ell-1}W_{V,\ell},
\end{equation}
\begin{equation}\label{eq:nn_att_E}
    E_{\ell} = (Q_{\ell}K_{\ell}^T)\circ \bsq{1/\sqrt{\abs{Q}}}_{225\times225} + H,
    \vspace{-1em}
\end{equation}
\begin{minipage}{0.30\textwidth}
\begin{equation}\label{eq:nn_att_P}
    P_{\ell} = {\tt Softmax} (E_{\ell}, \text{dim=1}),
\end{equation}
\end{minipage}
\begin{minipage}{0.18\textwidth}
\begin{equation}\label{eq:nn_att_Z}
    \bar{Z}_{\ell} = P_{\ell}V_{\ell},
\end{equation}
\end{minipage}\bigskip\newline
\end{subequations}
where the row vectors of matrices $Q_{\ell}$, $K_{\ell}$, $V_{\ell}$ are called query, key, and value vectors, respectively, learned from the corresponding row vectors in matrix $Z_{\ell-1}$; $\abs{Q}$ and $\abs{V}$ are the dimensions of the query and value vectors; $[x]_{a\times b}$ is a matrix of size $a\times b$ with all entries equal to $x$; $\circ$ is the element-wise Hadamard product; the element $e_{i,j}$ in $i$th row and $j$th column of $E_{\ell}$ represents a normalized similarity score induced by dot-product between $i$th query vector in $Q_{\ell}$ and $j$th key vector in $K_{\ell}$; $E_{\ell}$ essentially encodes how similar two row vectors in $Z_{\ell-1}$ are with each other; \eqref{eq:nn_att_P} apply ${\tt Softmax}(\cdot)$ to each column of $E_{\ell}$; the element $p_{i,j}$ in $i$th row and $j$th column of $P_{\ell}$ is equal to $\exp(e_{i,j})/\sum_k\exp(e_{k,j})$ such that each column vector of $P_{\ell}$ is a weight vector; each row vector in $\bar{Z}_{\ell}$ is a weighted summation of value vectors in $V_{\ell}$ using the weights learned in $P_{\ell}$. 

Notably, in the first layer $\ell=1$, the mask $H$ in \eqref{eq:nn_att_E} sets the similarities between appended zero row vectors and other row vectors in $Z_{0}=Z$ to $-\infty$. Subsequently, if the $i$th or $j$th row vector in $Z$ is an appended zero vector, \eqref{eq:nn_att_P} results in weights $p_{i,j}=0$ in $P_{1}$ and \eqref{eq:nn_att_Z} yields the $i$th or $j$th row also a zero vector in $\bar{Z}_{1}$. Furthermore, the computation in~\eqref{eq:nn_backbone} inherits the zero-row vectors from $\bar{Z}_{1}$ to $Z_{1}$. Inductively, through $\ell=1,2,3$, the attention backbone preserves the zero rows in $Z_0=Z$. Eventually, the attention backbone outputs $Z_3=[f^{i}_k, \dots, f^{j}_k, \dots]^T$ where each row vector $(f^{\iota}_k)^T$ is a learned feature vector corresponds to the input row vector $(z^{\iota}_k)^T$ in $Z=[z^{i}_k, \dots, z^{j}_k, \dots]^T$.

\subsection{Policy Head}\label{subsec:net_head}
Similar to~\eqref{eq:reward_R}, we use the SVO category $\sigma^i$ of the $i$th driver to combine the learned information $\{f^{j}_k\}_{j\in A(i)}$ from adjacent vehicles $j\in A(i)$ with $f^{i}_k$ of the driver $i$ and attain a single vector $\bar{f}^{i}_k\in\R^{225}$ according to,
\begin{equation}\label{eq:nn_head_svo}
    \bar{f}^{i}_k = \alpha(\sigma^i) f^{i}_k + \beta(\sigma^i)\Sigma_{j\in A(i)}f^{j}_k.
\end{equation} 
We use another MLP similar to~\eqref{eq:nn_mlp} with a bypass residual connection. This is followed by the final element-wise ${\tt Softmax}$ activation function that admits the following form,
\begin{equation}\label{eq:nn_softmax}
    {\tt Softmax}(z)=\exp(z_i)/\Sigma_i\exp(z_i),
\end{equation}
where $z$ is a column vector and $z_i$ is the $i$th element in $z$. The ${\tt Softmax}$ calculates a probability distribution as the policy distribution output $\pi_{NN}\in\R^{225}$.

The SANN architecture provides several advantages:
\begin{enumerate}
    \item The normalization process normalizes the input information using lane and vehicle dimensions which improves the prediction robustness to different highway structural dimensions and vehicle model types.
    \item The learning process is interaction-aware. The attention backbone interchanges information between each feature vector corresponding to each interacting driver, which captures the inter-traffic dependencies in the personal reward~\eqref{eq:reward_r}.
    \item The learning process is cooperation-aware. The policy head fuses the learned features using the driver's SVO $\sigma^i$. This process emulates~\eqref{eq:reward_R} and introduces the notion of cooperation/competition into learning.
    \item The SANN incorporates the behavioral model priors $\sigma^i,w^i$ that are later estimated by a Bayesian filter. This offers better online transferability to different drivers.
    \item The SANN is permutation invariant, namely, interchanging the order of the inputs in $\{\bar{s}_k^j\}_{j\in A(i)}$  will not alter the value of the learned features $\{f_k^j\}_{j\in A(i)}$ or affect the final policies $\pi_{NN}$. Given the interacting drivers $j\in A(i)$ geographically located in a 2D plane, the SANN learned policy should not be affected by the artificial information carried in the input order. 
    \item The SANN can handle variable numbers of inputs up to a maximum number $N_z$. This offers better transferability to different traffic conditions/densities.
\end{enumerate}
These properties might not necessarily be preserved by other networks, e.g., Graph Neural Networks~\cite{gcn}, the LSTM~\cite{lstm}. Then, we use this neural network behavioral model to predict the trajectories of interacting vehicles for the decision-making of our ego vehicle in the forced merging scenario.
\section{Decision-Making Under Cooperation Intent Uncertainty}\label{sec:methodCtrl}
We use the Bayesian filter to infer drivers' latent model parameters from observed traffic interactions (Sec.\ref{subsec:inference}). Then, based on the behavioral model, we incorporate the predictions of interacting drivers' behavior into a receding-horizon optimization-based controller to generate reference trajectories for the ego vehicle (Sec.\ref{subsec:control}). In Sec.\ref{subsec:search}, we use an interaction-guided decision tree search algorithm to solve this optimization problem and integrate it with the SANN for prediction. The learned SANN improves online prediction efficiency and while the tree-search algorithm offers a probabilistic safety guarantee and good scalability.

\subsection{Bayesian Inference of Latent Driving Intentions}\label{subsec:inference}
At each time step $k$, we assume the ego vehicle $0$ interacts with adjacent vehicle $i\in A(0)$ and can observe its nearby traffic state $s_k^i, \boldsymbol{s}_k^{-i}$. Assuming that the $i$th driver's decision-making process follows the policy~\eqref{eq:behavior_model_policy}, we need to infer the latent social psychological parameters $\sigma^i,w^i$ in order to predict the future behavior/trajectory of vehicle $i$. We assume observation history of traffic around vehicle $i$ is available,
\begin{equation}\label{eq:xi}
    \xi_k^i = [s_0^i, \boldsymbol{s}_0^{-i}, s_1^i, \boldsymbol{s}_1^{-i}, \dots, s_k^i, \boldsymbol{s}_k^{-i}],
\end{equation}
which contains the state $s_n^i$ of vehicle $i$ and the aggregated state $\boldsymbol{s}_n^{-i}$ of adjacent vehicles around vehicle $i$ for all time step $n=0,1,\dots,k$. Then, we use the following Bayesian filter to recursively estimate a posterior distribution of the latent parameters $\sigma^i,w^i$ from $\xi_k^i$.

\begin{prop}\label{prop:bayesian}
    Given a prior distribution $\prob{\sigma^i,w^i|\xi^i_{k}}$ and assume the unmodeled disturbance $\tilde{w}_k^i\sim\mathcal{N}(0,\Sigma)$ in~\eqref{eq:fEoM} follows a zero-mean Gaussian distribution, then the posterior distribution $\prob{\sigma^i,w^i|\xi^i_{k+1}}$ admits the following form,
    \begin{equation}\label{eq:bayesian}
    \begin{array}{l}    
        \prob{\sigma^i,w^i|\xi^i_{k+1}} = \frac{1}{N\brk{\xi^i_{k+1}}}\cdot \\
        D(s^i_{k+1},\sigma^i,w^i,s_k^i, \boldsymbol{s}_k^{-i})\cdot\prob{\sigma^i,w^i|\xi^i_{k}},
    \end{array}
    \end{equation}
    where $N\brk{\xi^i_{k+1}}$ is a normalization factor, and 
    \begin{equation}\label{eq:bayesian_D}
    \begin{array}{l}
         D(s^i_{k+1},\sigma^i,w^i,s_k^i, \boldsymbol{s}_k^{-i})=\\
        \sum_{\gamma(s^i_k)\in\Gamma(s^i_k)}\Big[
        \prob{\tilde{w}_k^i = s^i_{k+1} - \gamma_1^1(s^i_k)}\\
        \cdot
        \pi\brk{\gamma(s^i_k)|\sigma^i,w^i,s_k^i, \boldsymbol{s}_k^{-i}}\Big].
    \end{array}
    \end{equation}
\end{prop}
We note that~\eqref{eq:bayesian_D} represents a transition probability of a driver moving from $s^i_k$ to $s^i_{k+1}$ following the kinematics \eqref{eq:fEoM} and policy \eqref{eq:behavior_model_policy}. We initialize the Bayesian filter with a uniform distribution. Meanwhile, we can replace $\pi$ in~\eqref{eq:bayesian_D} with $\pi_{NN}$ for faster online Bayesian inference. We also provide a proof of the proposition in the following:
\begin{proof}
    We apply the Bayesian rule to rewrite the posterior distribution according to,
    \begin{equation*}
    \begin{array}{l}
        \prob{\sigma^i,w^i|\xi^i_{k+1}} 
        =\prob{\sigma^i,w^i| s_{k+1}^i,\boldsymbol{s}_{k+1}^{-i},\xi^i_k}
        \\
        = \frac{\prob{\boldsymbol{s}_{k+1}^{-i}|\xi_k^i}}{\prob{s_{k+1}^i,\boldsymbol{s}_{k+1}^{-i}|\xi_k^i}}
        \cdot
        \prob{s_{k+1}^i|\sigma^i,w^i,\xi^i_k}\cdot\prob{\sigma^i,w^i|\xi^i_k}
        \\
        = \frac{1}{N\brk{\xi^i_{k+1}}}\cdot 
        D(s^i_{k+1},\sigma^i,w^i,s_k^i, \boldsymbol{s}_k^{-i})\cdot\prob{\sigma^i,w^i|\xi^i_{k}},
    \end{array}
    \end{equation*}
    where we define $N\brk{\xi^i_{k+1}} = \frac{\prob{s_{k+1}^i,\boldsymbol{s}_{k+1}^{-i}|\xi_k^i}}{\prob{\boldsymbol{s}_{k+1}^{-i}|\xi_k^i}}$ and rewrite the transition probability,
    \begin{equation*}
        \begin{array}{l}        
        D(s^i_{k+1},\sigma^i,w^i,s_k^i, \boldsymbol{s}_k^{-i})=\prob{s_{k+1}^i|\sigma^i,w^i,\xi^i_k}
        \\
        =\sum_{\gamma(s^i_k)\in\Gamma(s^i_k)}\Big[
        \prob{\tilde{w}_k^i = s^i_{k+1} - \gamma_1^1(s^i_k)}\\
        \cdot
        \pi\brk{\gamma(s^i_k)|\sigma^i,w^i,s_k^i, \boldsymbol{s}_k^{-i}}\Big].
        \end{array}
    \vspace{-2em}
    \end{equation*}
\end{proof}
\subsection{Receding-horizon Optimization-based Control}\label{subsec:control}
Leveraging the posterior from~\eqref{eq:bayesian}, we use a receding-horizon optimization-based controller to incorporate the trajectory predictions~\eqref{eq:behavior_model_policy} of interacting vehicles $i\in A(0)$ and plan a reference trajectory for the ego vehicle according to
\begin{equation}\label{eq:ego_MPC}
\begin{array}{c}
     \gamma^*(s_k^0) = \argmax_{\gamma(s_k^0)\in \Gamma(s_k^0)} Q_0\brk{\boldsymbol{s}_k^{-0}, \gamma(s_k^0)}
\end{array}.
\end{equation}
Similar to Sec.~\ref{subsec:behaviorModel}, we compute the control signal $u_n^{0}=[a_n^{0},\delta_n^{0}]^T$ at each time step $n=k,\dots,k+N'$ to track this reference trajectory $\gamma^*(s_k^0)$ for one control sampling period $N'\Delta T$ sec. Then, we update the reference trajectory using \eqref{eq:ego_MPC} after $N'\Delta T$ sec. Meanwhile, the cumulative reward function $Q_0\brk{\boldsymbol{s}_k^{-0}, \gamma(s_k^0)}$ admits the following form,
\begin{equation}\label{eq:ego_Q}
\begin{array}{c}
     Q_0\Big(\boldsymbol{s}_k^{-0}, \gamma(s_k^0)\Big) 
     =
     \frac{1}{\abs{A(0)}}\sum_{i\in A(0)}\\
    \Big[
     \E_{\sigma^i,w^i\sim\prob{\sigma^i,w^i|\xi^i_k}}
     Q'_0\Big(\boldsymbol{s}_k^{-0}, \gamma(s_k^0)|\sigma^i,w^i\Big)
     \Big],
\end{array}
\end{equation}
where the function value $Q'_0$ is computed according to,
\begin{equation}\label{eq:ego_Q_prime}
\begin{array}{c}
     Q'_0\Big(\boldsymbol{s}_k^{-0}, \gamma(s_k^0)|\sigma^i,w^i\Big) 
    =
     \E_{\gamma(s_k^i)\sim\pi\brk{\cdot|\sigma^i,w^i,s_k^i, \boldsymbol{s}_k^{-i}}}
     \\
     \Big[
     \sum\limits_{n=0}^{{\scriptscriptstyle  \lfloor N/N'\rfloor}} r_0 
     \Big(
     \gamma{\scriptscriptstyle  _{nN'}^{(n+1)N'}}(s_k^{0}), \gamma{\scriptscriptstyle  _{nN'}^{(n+1)N'}}(s_k^{i}) 
     \Big)
     \Big],
\end{array}
\end{equation}
and the ego vehicle acts to minimize its traveling time and avoid collision, thereby the ego reward function $r_0$ attains the following form,
\begin{equation*}
r_0\brk{\gamma_{n_1}^{n_2}(s_k^{0}), \gamma_{n_1}^{n_2}(s_k^{i})} =\neg c\brk{\gamma_{n_1}^{n_2}(s_k^{0}), \gamma_{n_1}^{n_2}(s_k^{i})}\cdot \tau(s^0_{k+n_2}).
\end{equation*}

The value of reward $Q_0$ is the averaged cumulative reward over pairwise interactions with all vehicles $i\in A(0)$ using predictions of their future trajectories. Equation~\eqref{eq:ego_Q} defines the expectation of the reward function with respect to the behavioral model parameters $\sigma^i,w^i$, while \eqref{eq:ego_Q_prime} samples trajectory predictions $\gamma(s^i_k)$ of vehicle $i$ from the policy $\pi$ conditioned on $\sigma^i,w^i$. In~\eqref{eq:ego_Q_prime}, $\pi$ can be replaced by $\pi_{NN}$ learned by the SANN to speed up the computations. Nonetheless, solving the problem \eqref{eq:ego_MPC} requires an exhaustive search over the trajectory set $\Gamma(s_k^0)$ that can be computationally demanding. Instead, we can treat the trajectory set  $\Gamma(s_k^0)$ as a decision tree (see Fig.~\ref{fig:trajectorySets}) and a tree-search-based algorithm can be developed to improve the computation efficiency.

\subsection{Interaction-Guided Decision Tree Search}\label{subsec:search}

\begin{algorithm}[ht]
\caption{Interaction-Guided Decision Tree Search}
\label{al:treeSearch}
\begin{algorithmic}[1]
\Require {$\Gamma(s^0_k)$, $A(0)$, $\pi_{NN}$, and $\xi_k^i$, $\Gamma(s^i_k)$, $\prob{\sigma^i,w^i|\xi^i_{k-1}}$ for all $i\in A(0)$}

\State initialize $ Q_0\brk{\boldsymbol{s}_k^{-0}, \gamma(s_k^0)} =0$ for all $\gamma(s^0_k)\in\Gamma(s^0_k)$

\State $\prob{\sigma^i,w^i|\xi^i_k}\propto
\Big[
\sum_{\gamma(s^i_k)\in\Gamma(s^i_k)} 
\prob{\tilde{w}_k^i = s^i_{k+1} - \gamma_1^1(s^i_k)}\cdot\pi_{NN}\brk{\gamma(s^i_k)|\sigma^i,w^i,s_k^i, \boldsymbol{s}_k^{-i}}
\Big]
\cdot \prob{\sigma^i,w^i|\xi^i_{k-1}}$ for all $i\in A(0)$ \Comment{{\color{lightgray}\footnotesize Bayesian filter using SANN imitated behavioral model}}

\State $\prob{\gamma(s_k^i)|\xi_k^i} = \sum_{\sigma^i, w^i}\pi_{NN}\brk{\gamma(s_k^i)|\sigma^i,w^i,s_k^i, \boldsymbol{s}_k^{-i}}\cdot\prob{\sigma^i,w^i|\xi^i_k}$ for all $i\in A(0)$ \Comment{{\color{lightgray}\footnotesize interaction behavior prediction using SANN imitated behavioral model}}

\State sort $A(0)$ w.r.t. $\brk{(x^0_k-x^i_k)^2+(y^0_k-y^i_k)^2}^{1/2}$, $i\in A(0)$ in a descending order
\For{$i\in A(0)$} \Comment{{\color{lightgray}\footnotesize prioritize search over closer interactions}}
    \ParFor{$\gamma(s_k^{0})\in\Gamma(s_k^{0})$}\Comment{{\color{lightgray}\footnotesize parallel search}}
        \For{$n=0\rightarrow {\scriptstyle\lfloor N/N'\rfloor}$} \Comment{{\color{lightgray}\footnotesize over prediction horizons}}
        \State $c_n=\sum_{\gamma(s_k^i)\in \Gamma(s_k^i)} \prob{\gamma(s_k^i)|\xi_k^i}\cdot c\brk{\gamma{\scriptscriptstyle  _{nN'}^{(n+1)N'}}(s_k^{0}), \gamma{\scriptscriptstyle  _{nN'}^{(n+1)N'}}(s_k^{i})}$ \Comment{{\color{lightgray}\footnotesize probability of collision with $i$}}
        \If{$c_n > 0.5$} \Comment{{\color{lightgray}\footnotesize trim unsafe decision tree branch}}
        \State $\Gamma(s_k^0)\leftarrow \Gamma(s_k^0)\setminus \Gamma_{\text{unsafe}}(s_k^0)$, $\Gamma_{\text{unsafe}}(s_k^0):=\set{\gamma\in \Gamma(s_k^0)|\gamma{\scriptscriptstyle  _{nN'}^{(n+1)N'}}=\gamma{\scriptscriptstyle  _{nN'}^{(n+1)N'}}(s_k^0)}$, and terminate all parallel search branches in $\Gamma_{\text{unsafe}}(s_k^0)$
        \Else
        \State $Q_0\brk{\boldsymbol{s}_k^{-0}, \gamma(s_k^0)} \leftarrow Q_0\brk{\boldsymbol{s}_k^{-0}, \gamma(s_k^0)} + \lambda^{n
        }\cdot r_0 \brk{\gamma{\scriptscriptstyle  _{nN'}^{(n+1)N'}}(s_k^{0}), \gamma{\scriptscriptstyle  _{nN'}^{(n+1)N'}}(s_k^{i}) }\cdot \prob{\gamma(s^i_k)|\xi^i_k}$ \Comment{{\color{lightgray}\footnotesize update discounted cumulative reward }}
        \EndIf
        \EndFor
    \EndParFor
\EndFor
\State $\gamma^*(s_k^0) = \argmax_{\gamma(s_k^0)\in \Gamma(s_k^0)} Q_0\brk{\boldsymbol{s}_k^{-0}, \gamma(s_k^0)} $
\\\Return{$\gamma^*(s_k^0)$}
\end{algorithmic}
\end{algorithm}

For online deployment, we incorporate the SANN-imitated behavioral model $\pi_{NN}$ into a decision tree search Algorithm~\ref{al:treeSearch} to facilitate the Bayesian filtering, trajectory prediction of interacting vehicles, and decision-making of the ego vehicle. As shown in Fig.~\ref{fig:trajectorySets}, the trajectory set $\Gamma(s^0_k)$ can be viewed as a decision tree and is initiated from the current state $s^0_k$ as the tree root. Each trajectory $\gamma(s^0_k)$ is a branch in the decision tree $\Gamma(s^0_k)$, each state in a trajectory is a node, and each (directed) edge encodes reachability from one node to another in a certain trajectory/branch. Meanwhile, two distinct branches $\gamma^{(m_1)}(s^0_k)$, $\gamma^{(m_2)}(s^0_k)$ can share a same trajectory segments, i.e., $(\gamma^{(m_1)})_0^{n_1}(s^0_k) = (\gamma^{(m_2)})_0^{n_1}(s^0_k)=\gamma_0^{n_1}(s^0_k)$. Algorithm~\ref{al:treeSearch} searches over this decision tree, updates the cumulative reward~\eqref{eq:ego_Q} for each branch, and trims unsafe branches, thereby, improving the searching efficiency. For example in Fig.~\ref{fig:trajectorySets}b), the on-ramp ego vehicle is trying to merge where there is a following highway vehicle in grey. The normal lane-changing trajectories/branches share the same subset of initial trajectory segments $\gamma_0^{n_1}(s^0_k)$ and are highly likely to cause a collision with the highway vehicle. Therefore, we can trim all the lane change trajectories from the decision tree (shown by semi-transparent lines in Fig.~\ref{fig:trajectorySets}b) and terminate further searches along these unsafe branches.  

This process is formalized in Algorithm~\ref{al:treeSearch}. In lines 2 and 3, we use the SANN behavioral model $\pi_{NN}$ to update the posterior distribution in the Bayesian filter (see Proposition~\ref{prop:bayesian}) and predict trajectory distributions that are used to compute the reward \eqref{eq:ego_Q}. Since a closer interacting vehicle is more likely to collide with the ego vehicle in the near future, we rank the indices of the adjacent vehicles in $A(0)$ in descending order with respect to their Euclidean distances to our ego vehicle. The three for-loops in line 5,6,7 of Algorithm~\ref{al:treeSearch} search over interactions with different vehicles, branches, and prediction horizons of a branch, respectively. In lines 8-13, the sorted set $A(0)$ prioritizes collision checks with closer interacting vehicles and trims the unsafe branches as early as possible. We trim all branches with the common set of nodes $\gamma{\scriptscriptstyle  _{nN'}^{(n+1)N'}}(s_k^0)$ if the ego vehicle of trajectory segment $\gamma{\scriptscriptstyle  _{nN'}^{(n+1)N'}}(s_k^0)$ has a probability of collision with the $i$th vehicle higher than a threshold of 0.5. Otherwise, we will update the cumulative reward according to line 12 in Algorithm~\ref{al:treeSearch}. Eventually, in line 17, we solve~\eqref{eq:ego_MPC} by simply choosing the branch with the maximum cumulative reward. We also note that the three for-loops enable linear scalability of this algorithm with respect to both the number of interacting drivers in $A(0)$ and the number of prediction horizons $\lfloor N/N'\rfloor$.
\section{Simulation and Experimental Results}\label{sec:results}
We first present both qualitative and quantitative results of real-world trajectory prediction using our Behavioral Model and the Bayesian Filter in Sec.~\ref{subsec:result_trajPred}. We report the performance of the SANN imitation learning in Sec.~\ref{subsec:result_SANN}. We provide extensive evaluations of the proposed decision-making framework in forced merging tasks using our simulation environment (Sec.~\ref{subsec:result_sim}), the real-world High-D traffic dataset~\cite{highD} (Sec.~\ref{subsec:result_highD}), and the Carla simulator~\cite{carla} (Sec.~\ref{subsec:result_carla}). We note that we \textbf{do not} need to re-tune the model hyperparameters or re-train the SANN for different forced merging evaluation environments. The demonstration videos are available in \url{https://xiaolisean.github.io/publication/2023-10-31-TCST2024} 

\subsection{Real-world Trajectory Prediction}\label{subsec:result_trajPred}
\begin{figure*}[ht!]
    \centering
    \includegraphics[width=0.98\textwidth]{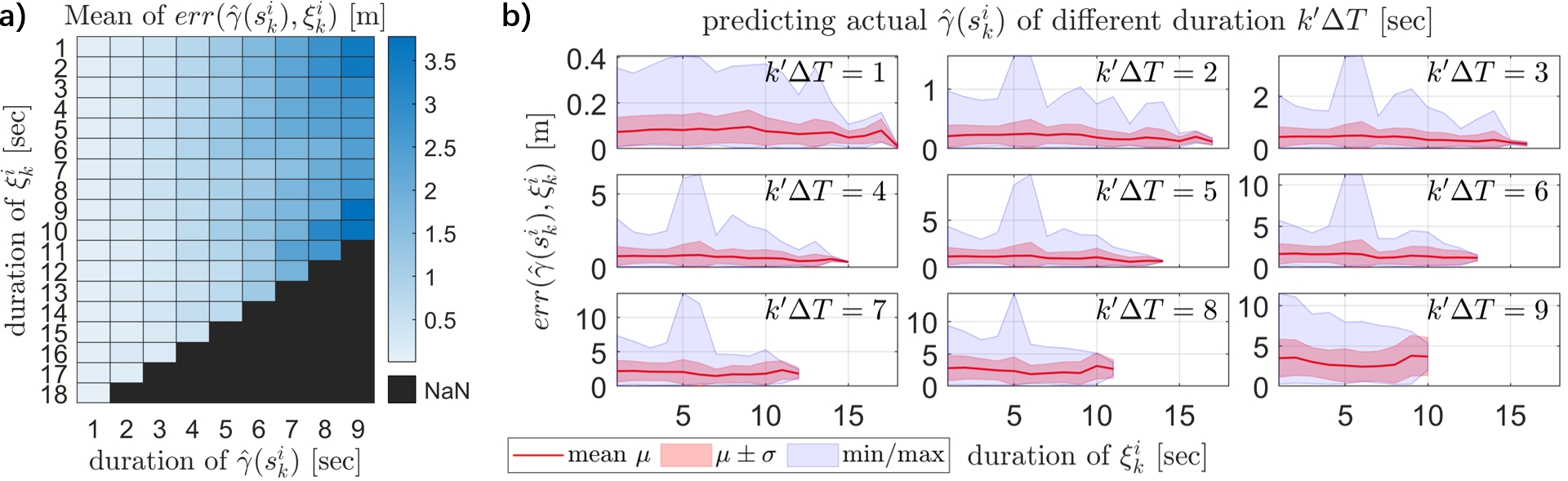}\vspace{-0.5em}
    \caption{Error statistics of trajectory prediction comprises 6,774 High-D trajectories of in total 28,078 driving seconds: (a) Each grid reports the mean of prediction errors using $\xi^i_k$, $\hat{\gamma}(s^i_k)$ of the same lengths. (b) Each subplot visualizes the mean prediction errors (red lines) corresponding to $\hat{\gamma}(s^i_k)$ of the same duration versus variable duration of $\xi^i_k$. We use red shaded areas to denote the one standard deviation and use blue shaded areas to represent the minimum/maximum error bounds.}
    \label{fig:filteringStats}
\end{figure*}

\begin{figure}[ht!]
    \centering
    \includegraphics[width=0.49\textwidth]{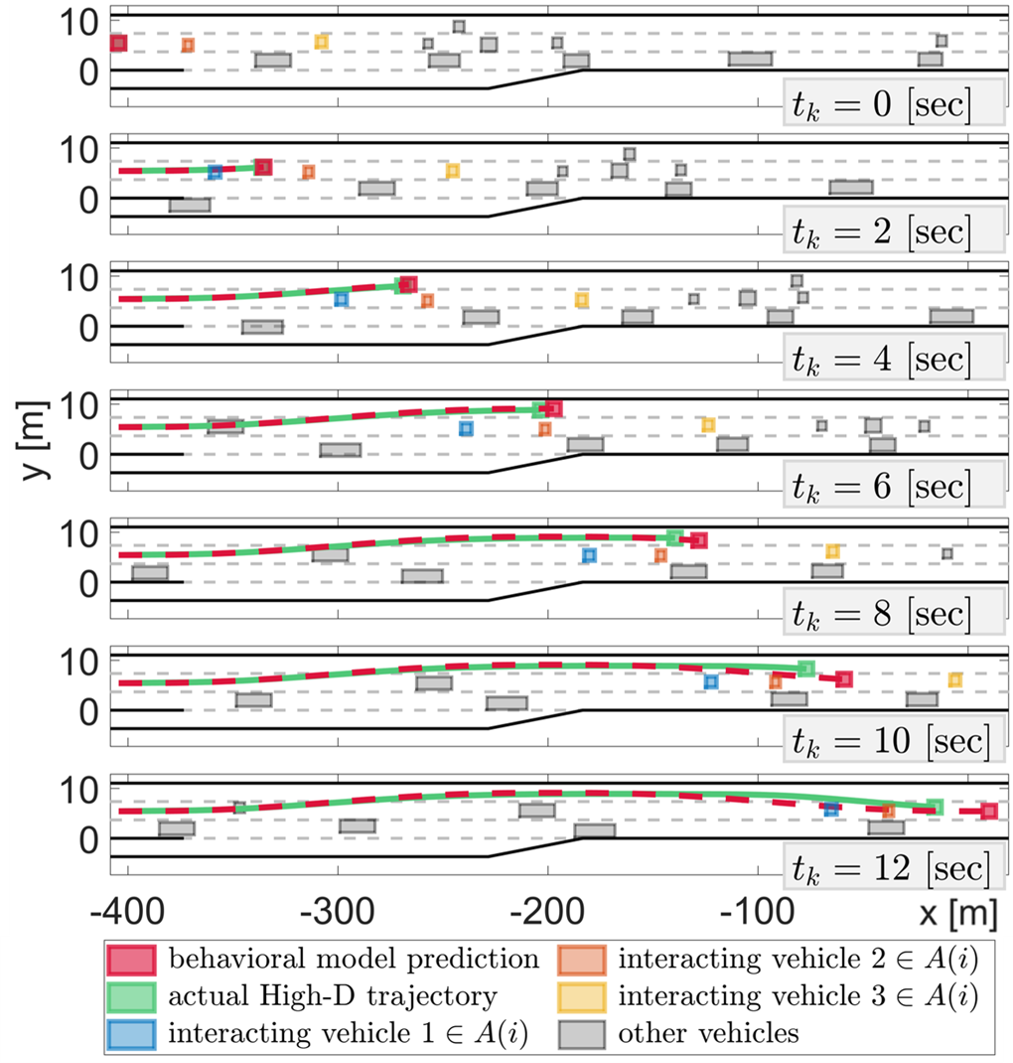}  
    \caption{Reproducing real-world overtaking trajectory: The trajectories of the virtual vehicle $i$ (in red) are synthesized using our behavioral model \eqref{eq:behavior_model_MPC} considering its driver of model parameters $\sigma^i=\text{``competitive"},$ $w^i=[0, 1, 0]^T$. This virtual ``competitive" driver overtakes the vehicle $2$, thereby, minimizing the traveling time $\tau$. The resulting trajectories in red match the actual trajectories in green.}
    \label{fig:overtakingPred}
\end{figure}

We present qualitative (Fig.~\ref{fig:overtakingPred}) and quantitative results (Fig.~\ref{fig:filteringStats}) of predicting real-world trajectories using our behavioral model and the Bayesian filter. In our first task, we aim to reproduce a real-world trajectory from a naturalistic High-D~\cite{highD} dataset. In a High-D traffic segment (Fig.~\ref{fig:overtakingPred}) of 12 sec, we identify a target vehicle $i$ (in green) that is overtaking its leading vehicle $2$ (in orange). We initialize a virtual vehicle (in red) at $t_k=0$ using the actual vehicle state $s_k^i$, and we set $\sigma^i=\text{``competitive"},$ $w^i=[0, 1, 0]^T$ in the behavioral model~\eqref{eq:behavior_model_MPC}. Afterwards, we control this virtual vehicle using~\eqref{eq:behavior_model_MPC} assuming no tracking error, and we set the control sampling period to $N'\Delta T=0.5\;\rm sec$. We compare the synthesized trajectories using our behavioral model with the actual ones in Fig.~\ref{fig:overtakingPred}. Our behavioral model demonstrates a good prediction accuracy from 0-6 sec, which is adequate for a predictive controller with a sampling period $N'\Delta T\leq 6\;\rm sec$. We also note that the prediction error is large at $t_k=12\;\rm sec$ because the error accumulates over longer prediction windows. Our prediction captures the overtaking trajectories and qualitatively demonstrates the effectiveness of our method in modeling real-world driving behavior.

Fig.~\ref{fig:filteringStats} summarizes the error statistics of the trajectory prediction. In online application scenarios, given an observed interacting history $\xi^i_{k}$, we recursively infer the latent behavioral model parameters $\sigma^i,w^i$ using the Bayesian filter~\eqref{eq:bayesian} as a posterior distribution $\prob{\sigma^i,w^i|\xi^i_{k}}$. Subsequently, the interacting vehicles' trajectories are predicted as a distribution using policy~\eqref{eq:behavior_model_policy} according to $\PP\brk{\gamma(s_k^i)|\xi_k^i} = \sum_{\sigma^i, w^i}\pi\brk{\gamma(s_k^i)|\sigma^i,w^i,s_k^i, \boldsymbol{s}_k^{-i}}\cdot\prob{\sigma^i,w^i|\xi^i_k}$. We quantify the prediction error between the actual trajectory $\hat{\gamma}(s_k^i)=\set{\hat{s}_{n}^{i}}_{n}^{k+k'}$ and the predicted trajectory distribution $\prob{\gamma(s_k^i)|\xi_k^i}$ using the following metric
\begin{equation}\label{eq:pred_err}
\begin{array}{l}
    {\tt err} \brk{\hat{\gamma}(s_k^i), \xi^i_k} = \E_{\gamma(s_k^i)\sim\prob{\gamma(s_k^i)|\xi_k^i}}\\
    \bsq{\frac{1}{k'+1}\sum\limits_{s_n^i\in\gamma(s_k^i), n=k
    }^{n=k+k'} \norm{\begin{bmatrix}x^i_n-\hat{x}^i_n\\y^i_n-\hat{y}^i_n\end{bmatrix}}_2 },
\end{array}
\end{equation}
where $k'\Delta T$ in second is the duration of the actual trajectory $\hat{\gamma}(s_k^i)$. This metric computes the expected $\ell_2-$norm error in position prediction averaged over time steps.

\begin{figure*}[ht!]
    \centering
    \includegraphics[width=0.98\textwidth]{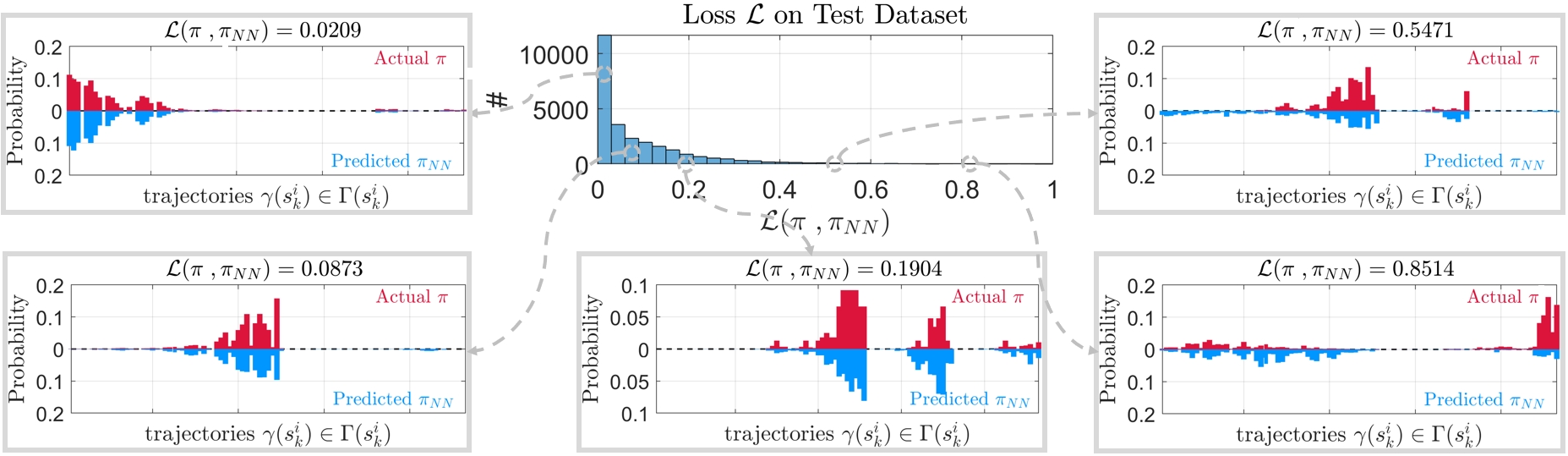}\vspace{-0.5em}
    \caption{Test statistics and examples of imitation learning on the test dataset. The histogram presents the loss statistics between the SANN learned policy distributions $\pi_{NN}$ and the actual ones $\pi$ computed from the behavioral model~\eqref{eq:behavior_model_policy}. Five qualitative examples are visualized in call-out boxes: The $y$-axis shows the probability of driver $i$ taking a certain trajectory $\gamma(s^i_k)$. For comparison, policies $\pi$ and $\pi_{NN}$ are plotted above and below the dashed line, respectively, in mirror reflection}
    \label{fig:nnResults}
\end{figure*}

We sample different traffic segments of different duration from the High-D dataset, and each traffic segment is bisected by time instance $t_k$ into training segments $\xi_k^i$ and prediction segment $\hat{\gamma}(s_k^i)$ corresponding to a sampled vehicle $i$. We apply the aforementioned procedure to each training segment $\xi_k^i$ and calculate the prediction error~\eqref{eq:pred_err} using the corresponding prediction segment $\hat{\gamma}(s_k^i)$. Meanwhile, in the sequel, we assume $w^i$ in a finite set 
\begin{equation}\label{eq:W_set}
W = \set{
\begin{aligned}    
     [0,0,1],\; & [0,1,1]/2,\;   [0,1,0],\; [1,1,1]/3,\\
     [1,0,1]/2,\; & [1,1,0]/2,\; [1,0,0]
\end{aligned}
}
\end{equation}
to reduce the dimension of parameter space $(\sigma^i,w^i)$. Subsequently, we have 22 possible combinations of $(\sigma^i,w^i)$: We assign 7 different $w^i\in W$ to three $\sigma^i\neq${ ``altruistic"}; if $\sigma^i=${ ``altruistic"}, the weights $w^i$ do not matter for the altruistic driver as defined in~\eqref{eq:reward_R},~\eqref{eq:sigma}.

As shown in Fig.~\ref{fig:filteringStats}a), the mean prediction errors are below $4\;\rm m$ for predictions of 1-9 sec ahead and using training segments of 1-18 sec. We also note that longer training segments $\xi_k^i$ (see Fig.~\ref{fig:filteringStats}b) reduce both the prediction error standard deviation and its maximum values. However, for longer prediction windows of 7-9 sec (see Fig.~\ref{fig:filteringStats}b), the error accumulates and leads to larger standard deviations and mean errors which are also observed in Fig.~\ref{fig:overtakingPred}. For shorter prediction windows of 1-6 sec, we have a maximum error below $5\;\rm m$ for most of the cases and the standard deviation is smaller than $1.5\;\rm m$ (see Fig.~\ref{fig:filteringStats}b). The results in Fig.~\ref{fig:overtakingPred} and Fig.~\ref{fig:filteringStats} provide evidence that our algorithms have good quantitative accuracy over shorter prediction windows, and good qualitative accuracy over longer prediction windows. Based on these considerations, we set the trajectory length as $N\Delta T= 6\;\rm sec$, so that we have a good prediction performance over a shorter prediction window of 6 sec. This duration covers a complete lane change of $T_{\text{lane}}= 4\;\rm sec$, and suffice the task of predicting interacting vehicles' trajectories for ego vehicle control. 

\subsection{Imitation Learning with SANN}\label{subsec:result_SANN}
The goal of imitation learning is to train the SANN to mimic the behavioral model. Namely, the predicted policy $\pi_{\text{NN}}$ should match the actual one $\pi$ from behavioral model~\eqref{eq:behavior_model_policy} accurately. We leverage the High-D dataset~\cite{highD} to synthesize realistic traffic $s_k^i, \boldsymbol{s}_k^{-i}$ for training the SANN. We randomly sample a frame from the High-D traffic together with a target vehicle $i$, thereby we can extract states $s_k^i, \boldsymbol{s}_k^{-i}$ of vehicles $i$ and its interacting vehicles. Together with sampled parameters $\sigma^i,w^i$, we can compute the actual policy distribution $\pi\brk{\cdot|\sigma^i,w^i,s_k^i, \boldsymbol{s}_k^{-i}}$ using~\eqref{eq:behavior_model_policy}. We repeat this procedure to collect a dataset of 183,679 data points. We decompose the dataset into training (70\%), validation (15\%), and test datasets (15\%). The MLP in the Attention Backbone has three layers of sizes 7, 32, 225, respectively. The MLP in the Policy Head has two layers of sizes 225, 225. Furthermore, we set $\abs{Q}=32$, $\abs{V}=225$, and the maximum number of feature vectors as $N_z=9$.

We use Python with Pytorch~\cite{pytorch} to train the SANN using the training dataset and the loss function~\eqref{eq:kl_loss}. We train the SANN for 1000 epochs using a Batch Stochastic Gradient Descent algorithm with momentum (batch size of 200). We set the initial learning rate and momentum to $0.01$ and $0.9$, respectively. We also adopt a learning rate scheduler that decreases the learning rate by half every 200 epochs. The training process takes in total 7 hours (25 sec per epoch) on a computer with 16 GB RAM and Intel Xeon CPU E3-1264 V3. We evaluate the performance of the SANN in the task of predicting the policy distributions in the test dataset, and the results are reported in Fig.~\ref{fig:nnResults}. For the majority of the test cases, the SANN achieves losses smaller than 0.4 which corresponds to a good statistical performance. Moreover, in the example (see Fig.~\ref{fig:nnResults}) with relatively larger losses of 0.8514, the learned policy (in blue) also qualitatively matches the actual one (in red). To conclude, the SANN imitates the proposed behavioral model with good accuracy. Hence, as presented in Algorithm~\ref{al:treeSearch}, we use this SANN learned policy to perform Bayesian filtering and trajectory prediction to facilitate the online decision-making for our ego vehicle.

\subsection{Forced Merging in Simulations}\label{subsec:result_sim}
\begin{figure*}[ht!]
    \centering  
    \includegraphics[width=0.98\textwidth]{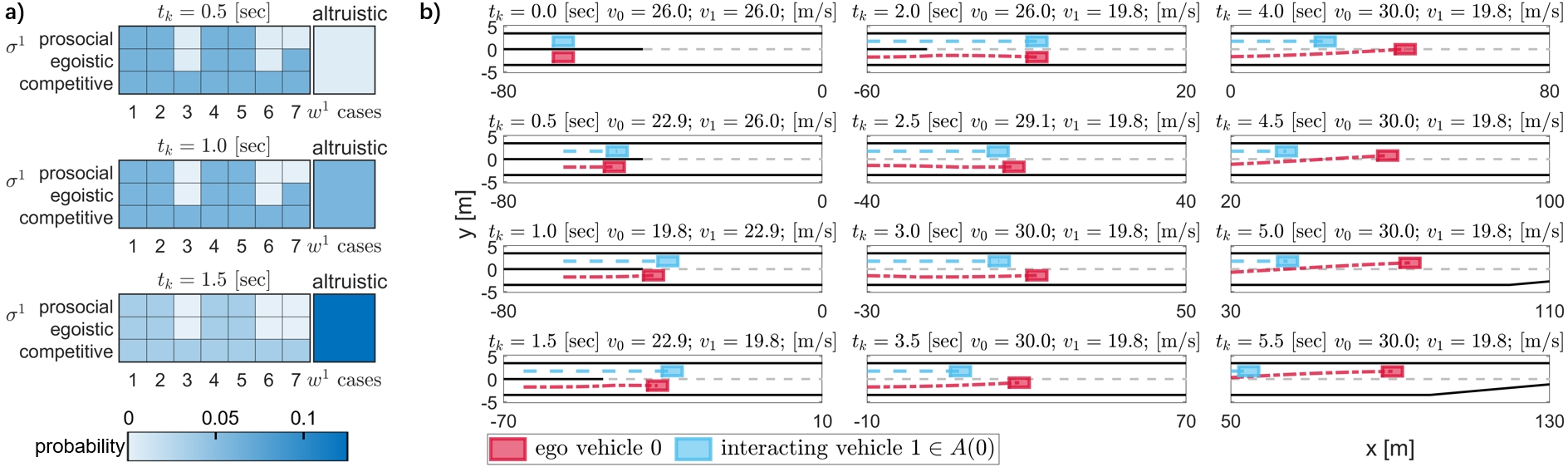}\vspace{-0.5em}
    \caption{A forced merging example of proactive interaction in a simulation: (a) Each subplot reports a posterior distribution of $\prob{\sigma^i,w^i|\xi^i_k}$ at a certain time instance $t_k$ from the Bayesian filter~\eqref{eq:bayesian}. The $x$-axis shows 7 cases of $w^1\in W$, and the $y$-axis shows three different SVO categories with $\sigma^i=${ ``altruistic"} stands along; (b) Each subplot visualizes a frame of highway interaction between the ego vehicle and vehicle 1 controlled by Algorithm~\ref{al:treeSearch} and behavioral model~\eqref{eq:behavior_model_MPC}, respectively.}
    \label{fig:forcedMergingSim}
\end{figure*}

We set up a simulation environment (see Fig.~\ref{fig:forcedMergingSim}) where the interacting vehicle is controlled by the behavioral model~\eqref{eq:behavior_model_MPC} with $\sigma^1=\;${``altruistic"}. We use the proposed Algorithm~\ref{al:treeSearch} to control the ego vehicle. For this and the following experiments, we set the control sampling period to $N'\Delta T=0.5\;\rm sec$ for both Algorithm~\ref{al:treeSearch} and the behavioral model. Instead of passively inferring the driver’s intention from its behavior, the Algorithm~\ref{al:treeSearch} controls the ego vehicle and exhibits a merging strategy with proactive interaction to test the interacting driver’s intention.

Specifically, in Fig.~\ref{fig:forcedMergingSim}b), the interacting driver 1 first drives at a constant speed from $0$ to $0.5$ sec. Meanwhile, from $0$ to $1$ sec, the ego vehicle 0 is not certain if driver 1 will yield the right of way for its merging attempt and, therefore it tentatively merges after vehicle 1 with longitudinal deceleration. Meanwhile, in the time interval between $0.5$ and $1.5$ sec, the ``altruistic" driver of vehicle 1 notices the merging intention of the ego vehicle and decelerates to promote merging. In the aforementioned interactions, the ego vehicle gradually updates its belief of the latent model parameters $\sigma^1,w^1$ of driver 1. As shown in Fig.~\ref{fig:forcedMergingSim}a), our algorithm correctly infers the ``altruistic" identity of driver 1. Subsequently, the ego vehicle aborts the merging action due to safety concerns with longitudinal acceleration to build up speed advantage for future merging. Then, in the time interval between $2.0$ and $5.5$ sec, being aware of the yielding behavior, the ego vehicle re-merges before vehicle 1 with longitudinal acceleration. This simulation example provides evidence that our method can effectively interpret the driving intentions of other drivers. Moreover, our decision-making module can leverage this information to facilitate the forced merging task while ensuring the safety of the ego vehicle.

\subsection{Forced Merging in Real-world Traffic Dataset}\label{subsec:result_highD}

\begin{figure}[ht!]
    \centering
    \includegraphics[width=0.49\textwidth]{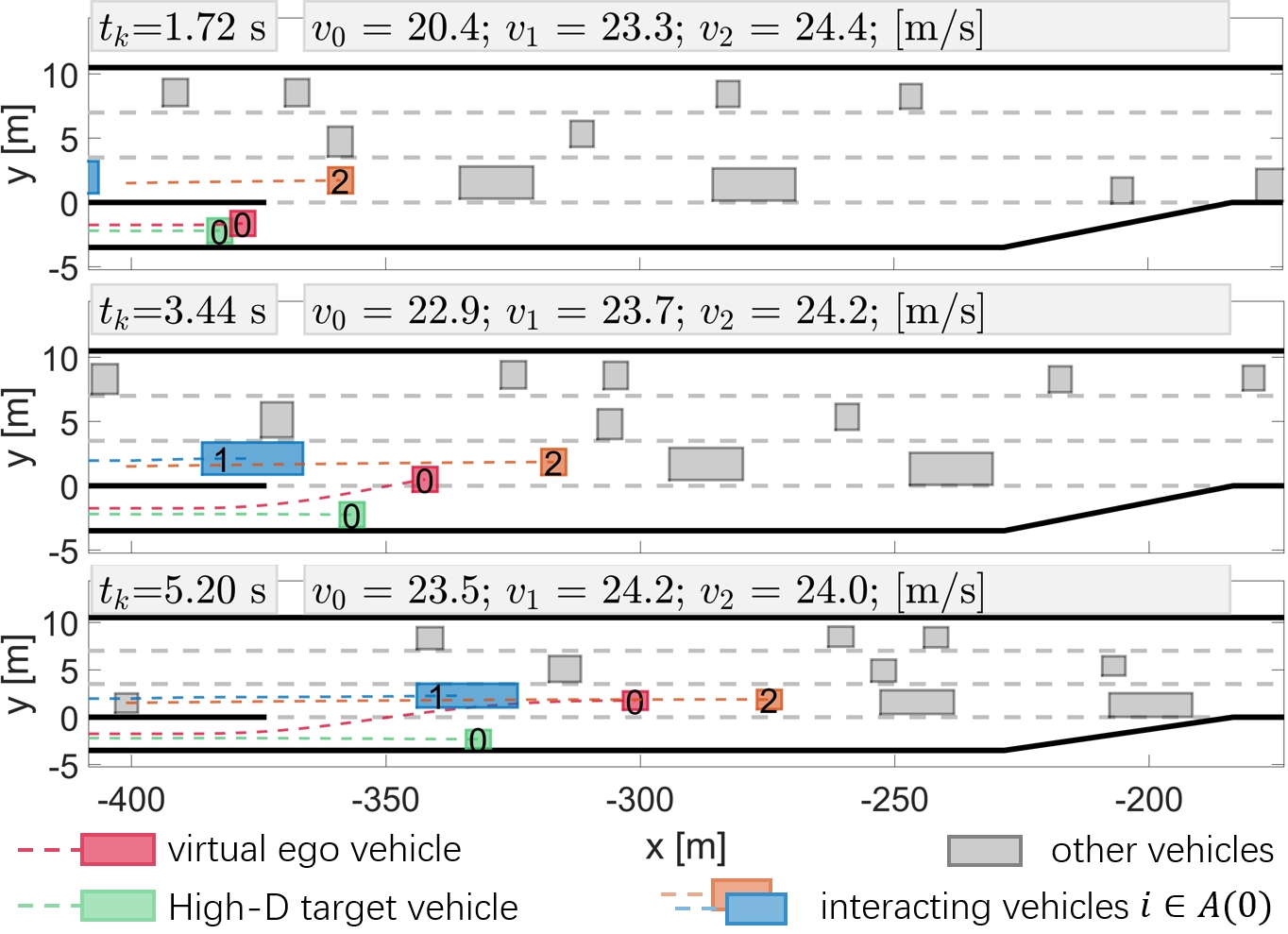}  
    \caption{A forced merging evaluation example on the High-D dataset: Our ego vehicle (in red) accelerates first to create sufficient gaps between highway vehicles 1 and 2, which are driving approximately at constant speeds. The ego vehicle successfully merges into the highway in $5.2\;\rm sec$ which is faster than the actual human driver (in green).} 
    \label{fig:highDEx1}
\end{figure}

\begin{figure}[ht!]
    \centering
    \includegraphics[width=0.49\textwidth]{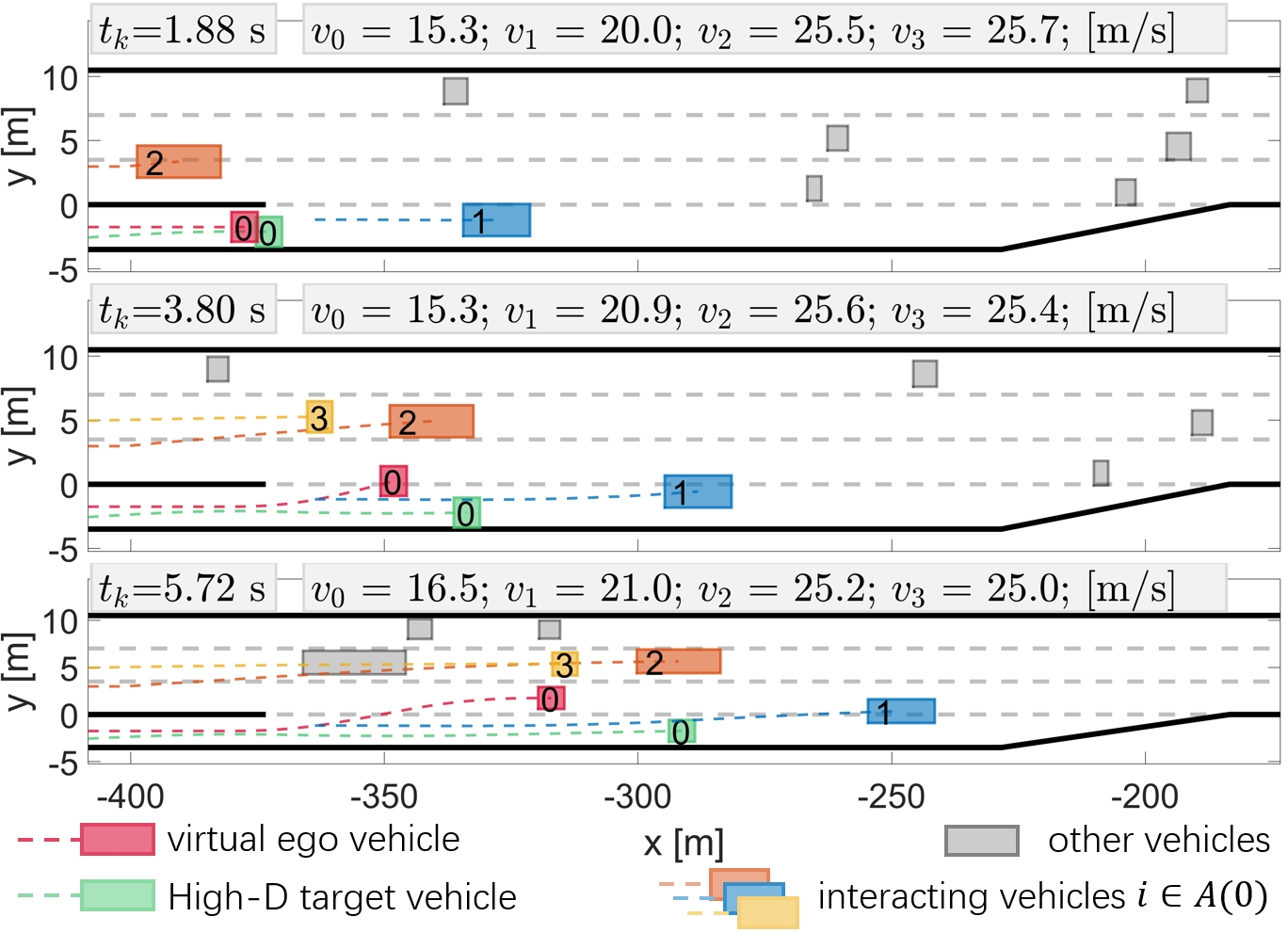}  
    \caption{Another forced merging evaluation example on the High-D dataset: The interacting vehicle 2 changes its lane in order to promote the merging action of the on-ramp vehicle. Our ego vehicle (in red) merges into the highway once observing the lane change behavior of vehicle 2. The ego vehicle successfully merges into the highway in $5.72\;\rm sec$ which is faster than the actual human driver (in green).}
    \label{fig:highDEx2}
\end{figure}

We further evaluate the performance of our method in the High-D~\cite{highD} real-world traffic dataset. There are 60 recordings in the High-D dataset where the recordings 58-60 correspond to highways with ramps. In recordings 58-60, we identify in total 75 on-ramp vehicles (High-D target vehicles) that merge into highways. For each one of the 75 vehicles, we extract its initial state at a recording frame when it appears on the ramp. Then, we initialize a virtual ego vehicle using this state and control this virtual ego vehicle using our decision-making module. Other vehicles are simulated using the actual traffic recordings, where we neglect the interaction between the virtual ego vehicle and the High-D target vehicle. Eventually, we generate 75 forced merging test cases repeating this procedure over all target vehicles. We visualize two test cases out of 75 as examples in Figs.~\ref{fig:highDEx1} and~\ref{fig:highDEx2}. The two interacting vehicles 1 and 2 in Fig.~\ref{fig:highDEx1} are approximately driving at constant speeds. Therefore, our decision-making module accelerates the virtual ego vehicle to a comparable speed and merges the virtual ego vehicle in between the two interacting vehicles. In the second example (see Fig.~\ref{fig:highDEx2}), our decision-making module can interpret the yielding behavior of vehicle 2 and, thereby merge the ego vehicle highway in a timely manner. In both examples, our algorithm is able to complete the forced-merging task faster than the real-world human drivers (in green boxes). 

\begin{figure}[ht!]
    \centering
    \includegraphics[width=0.49\textwidth]{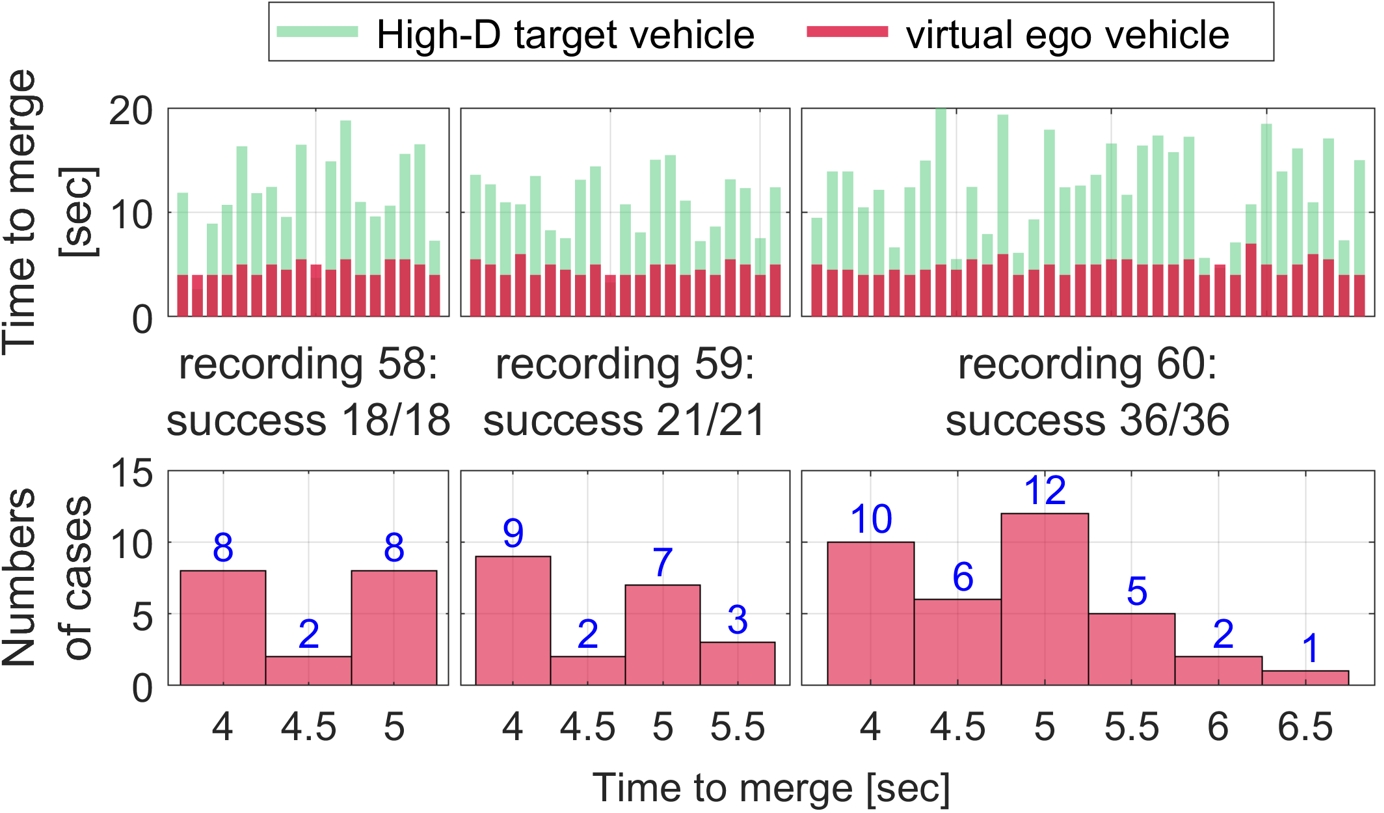}  
    \caption{Forced merging test statistics in High-D: The results are reported separately for the three recordings, and our method achieves a 100\% success rate. (Upper) We visualize the time-to-merge of the virtual ego vehicles (red bars) in comparison with the ones of the actual High-D vehicles (green bars). (Lower) Three histograms visualize the number of cases in which our method takes certain seconds to merge the virtual ego vehicle.}
    \label{fig:highDTime}
\end{figure}

Meanwhile, we identify a test case failure if our method fails to merge the virtual ego vehicle into the highway before the end of the ramp, or if the ego vehicle collides with other vehicles and road boundaries. Otherwise, we consider it a success case. In the three recordings, we have 18, 21, and 36 test cases, respectively. Our method can achieve a success rate of 100\% in all recordings. We also report the results of the time it took to merge the virtual ego vehicle in Fig.~\ref{fig:highDTime}. Compared to the actual human driver of the target vehicle (green bars), our decision-making module can merge the ego vehicle into the highway using a shorter time. Meanwhile, for 64 out of the 75 test cases, our method completes the task within $5\;\rm sec$ while we model a complete lane change with $T_{\text{lane}}=4\;\rm sec$ in Sec.~\ref{subsec:lanechange}. The results demonstrate that our method can successfully complete the forced merging task with traffic in real-world recordings in a timely and safe manner. Though the traffic is realistic in the dataset, we also note that there is no actual interaction between the virtual ego vehicle and other vehicles. Namely, the recorded vehicles in the dataset will not respond to the action of our virtual ego vehicle. 

\begin{figure*}[ht!]
    \centering  
    \includegraphics[width=0.98\textwidth]{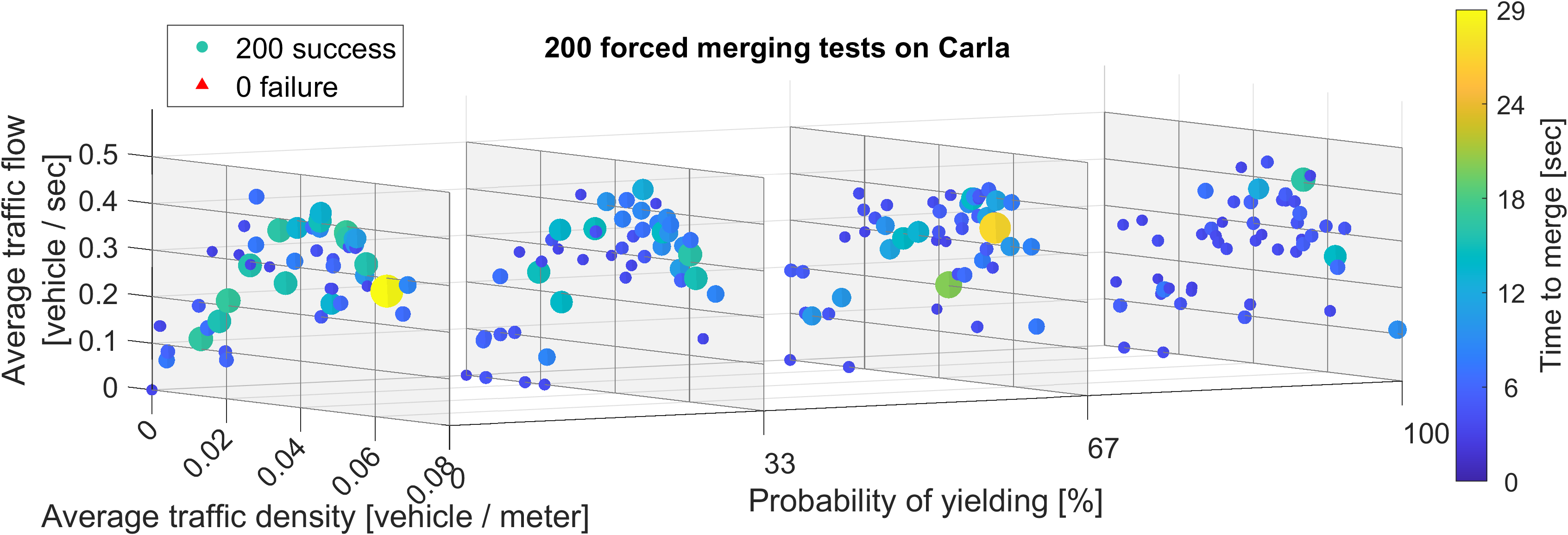}
    \caption{Forced merging test statistics in the Carla simulator with different traffic conditions: The $x$-axis shows four different traffic settings where vehicles have different probabilities of being willing to yield to the ego vehicle. We have 50 test cases for each of the four traffic settings. Each dot is a single test case where the color and the size of the dot reflect the time needed to merge the ego into the highway. The $y$ and $z$ axes report the average density and flow of the traffic in the ROI.} 
    \label{fig:carlaStats}
\end{figure*}

\begin{figure}[ht!]
    \centering
    \includegraphics[width=0.49\textwidth]{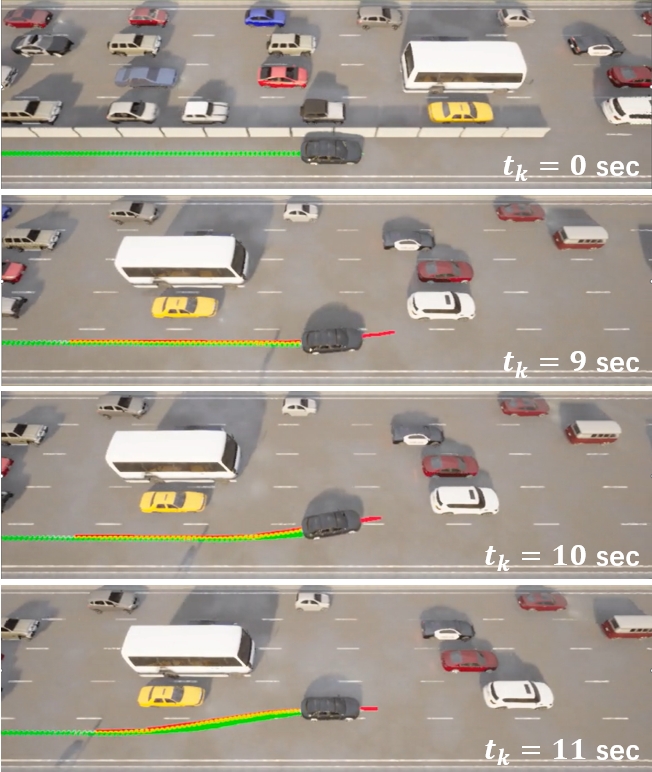}  
    \caption{A forced merging example in the Carla simulator: the ego vehicle is the black vehicle in the middle lower side of each subplot with its trajectories in green lines and reference trajectories (from our decision-making module) in red lines.}
    \label{fig:carlaDemo}
\end{figure}
\subsection{Forced Merging in Diverse Carla Traffics}\label{subsec:result_carla}
We set up a forced merging scenario in the Carla simulator~\cite{carla} (see Fig.~\ref{fig:carlaDemo}) and test our decision-making module in diverse and reactive simulated traffic. The vehicles in the traffic are controlled by the default Carla Traffic Manager. In the previous experiments, we assumed that the ego vehicle was able to accurately track the reference trajectories $\gamma^*(s_k^0)$ from Algorithm~\ref{al:treeSearch}. To mimic the real application scenario of our algorithm, we developed a PID controller for reference trajectory tracking. As visualized in the system diagram of Fig.~\ref{fig:systemArch}, Algorithm~\ref{al:treeSearch} in the high-level decision-making module updates optimal reference trajectories $\gamma^*(s_k^0)$ every $N'\Delta T=0.5\;\rm sec$. Meanwhile, to track this reference trajectory $\gamma^*(s_k^0)$, the low-level PID controller computes the steering and throttle signal of the actual vehicle plant (simulated by Carla using Unreal Engine) at 20 $\rm Hz$. 

\begin{figure}[ht!]
    \centering
    \includegraphics[width=0.49\textwidth]{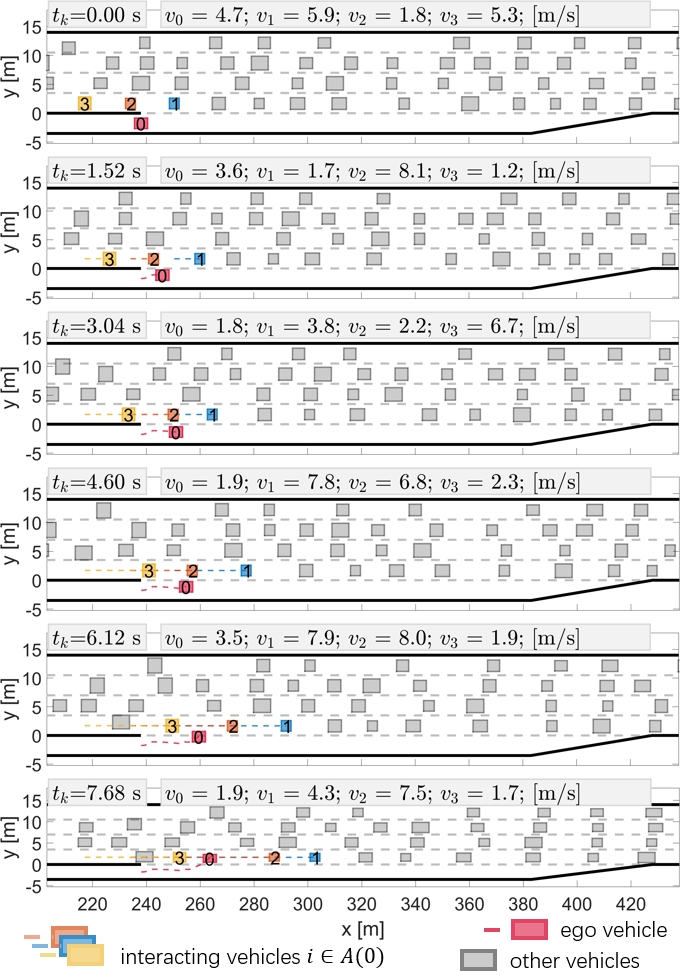}  
    \caption{A forced merging example in a dense Carla platoon: we use the default Carla Traffic Manager to control the highway vehicles. Our ego vehicle actively searches for gaps in the platoon and successfully merges into this dense traffic. $v_0\leq v_{\min}$ is due to the tracking error of the PID controller.}
    \label{fig:carlaEx1}
\end{figure}

Fig.~\ref{fig:carlaEx1} illustrates our method capability to merge the ego vehicle into a dense highway platoon in Carla. From $0$ to $4.60\;\rm sec$, the ego vehicle attempts to merge between vehicles 1 and 2 and, eventually aborts lane change due to safety concerns. From $4.60$ to $6.12\;\rm sec$, vehicle 3 is decelerating due to a small headway distance. And vehicle 2 is accelerating seeing an enlarged headway distance as a result of the acceleration of vehicle 1 before $4.60\;\rm sec$. Consequently, the gap between vehicles 2 and 3 is enlarged due to this series of interactions from $4.60$ to $6.12\;\rm sec$. Taking advantage of these interactions, our ego vehicle decides to accelerate and merge before vehicle $3$ at $4.60\;\rm sec$. 

Moreover, we are interested in quantifying the state of the traffic that the ego vehicle directly interacts with, i.e., the traffic in the lane near the ramp. We first define the traffic Region Of Interest (ROI) as the traffic in the near-ramp lane between the longitudinal position $x_{\text{ramp}}-l_{\text{ramp}}$ and $x_{\text{ramp}}$ (see Fig.~\ref{fig:forcedMergingProblem}). Subsequently, we define the following two variables: We compute the average traffic flow as the number of vehicles entering this traffic ROI over the entire simulation time; and we calculate the traffic density as the number of vehicles located in this traffic ROI over the length of this region $l_{\text{ramp}}$ at a certain time instance. Then, we average the traffic density over measurements taken every $\Delta T=0.05\;\rm sec$ which yields the average traffic density. 

The two variables quantify the averaged speed and density of the near-ramp traffic that the ego vehicle directly interacts with. For the example in Fig.~\ref{fig:carlaEx1}, the traffic in the ROI has an average traffic density of $0.055\;\rm vehicle/meter$ and an average traffic flow $0.3646\;\rm vehicle/sec$. We also note that the vehicles controlled by the default Carla Traffic Manager have no interpretation of other drivers’ intentions and, therefore, are unlikely to yield to the ego vehicle. This introduces more difficulties in merging into a dense vehicle platoon (such as the example in Fig.~\ref{fig:carlaEx1}) even for human drivers. 

Therefore, we introduce another variable, i.e., the probability of yielding, to have more diversified evaluations for our method. Specifically, for traffic with zero probability of yielding, we control all vehicles in the traffic using the default Carla Traffic Manager. For that being set to a non-zero value of $X\%$, the vehicles in the traffic ROI have $X\%$ probability of yielding to the on-ramp ego vehicle. The yielding behavior is generated by the Cooperate Forced Controller~\cite{kong2023simulation}. Similar to the Intelligent Driver Model (IDM)~\cite{idm} that controls a vehicle to follow the leading vehicle ahead, the Cooperate Forced Controller is a car-following controller derived from the IDM, but assuming the on-ramp ego vehicle is the leading vehicle.

We test our algorithm at various traffic conditions with different settings. As shown in Fig.~\ref{fig:carlaStats}, we achieve a 100 \% success rate among 200 test cases where we have different traffic densities and different probabilities of yielding for interacting vehicles. In the majority of the test cases, the ego vehicle takes less than  $9\;\rm sec$ to merge. Generally, in denser traffic, our algorithm needs more time to interact with more vehicles and, thus, more time to merge. With a higher probability of yielding, the algorithm takes less time to complete the merging tasks on average. Notably, our algorithm can achieve a 100\% success rate even in traffic with zero probability of yielding.
 
\section{Conclusions}\label{sec:conclusion}
In this paper, we proposed an interaction-aware decision-making module for autonomous vehicle motion planning and control.
In our approach, interacting drivers' intentions are modeled as hidden parameters in the proposed behavioral model
and are estimated using a Bayesian filter. 
Subsequently, interacting vehicles' behaviors are predicted using the proposed behavioral model. 
For the online implementation, the Social-Attention Neural Network (SANN) was designed and utilized to imitate the behavioral model in predicting the interacting drivers' behavior.  The proposed approach is easily transferable to different traffic conditions. 
In addition, the decision tree search algorithm was proposed for faster online computation; this algorithm leads to linear scalability with respect to the number of interacting drivers and prediction horizons. 
Finally, a series of studies, based on naturalistic traffic data and simulations, were performed to demonstrate the capability of the proposed decision-making module.  In particular, we have shown that the behavioral model has good prediction accuracy, and the proposed algorithm is successful in merging the ego vehicle in various simulations and real-world traffic scenarios, without the need for returning the model hyperparameters or retraining the SANN.



\bibliographystyle{IEEEtran}
\bibliography{ref.bib}
\vspace{-4em}
\begin{IEEEbiography}[{\includegraphics[width=1in,height=1.25in,clip,keepaspectratio]{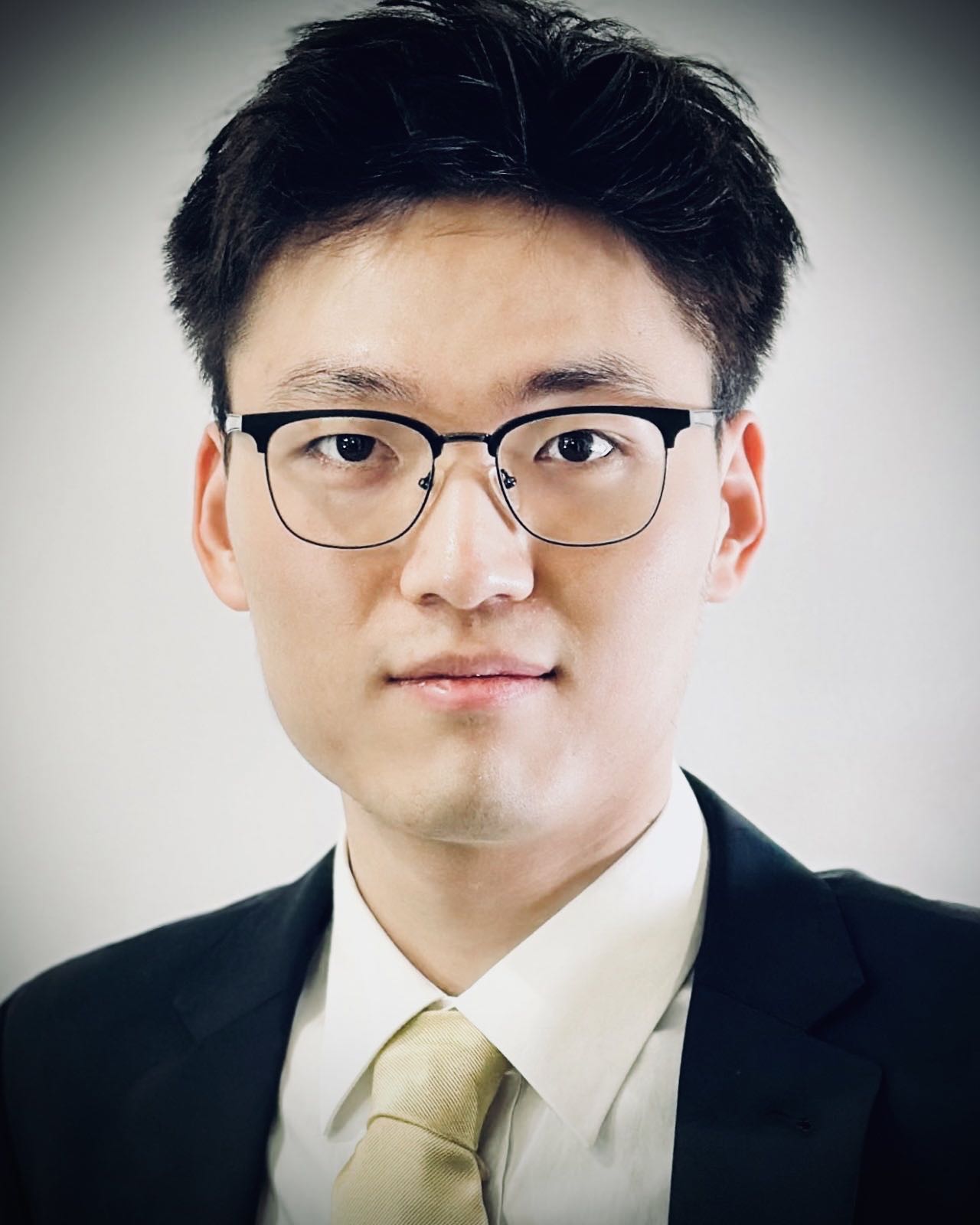}}]{Xiao Li}
received the B.S. degree in mechanical engineering from Shanghai Jiao Tong University, Shanghai, China, in 2019, and the M.S. degree in mechanical engineering from the University of Michigan, Ann Arbor, MI, USA, in 2021, where he is currently pursuing Ph.D. degree in aerospace engineering. His research interests include learning-based methods in constrained optimization and in human-in-the-loop decision-making for autonomous agents.
\end{IEEEbiography}

\vspace{-4em}
\begin{IEEEbiography}[{\includegraphics[width=1in,height=1.25in,clip,keepaspectratio]{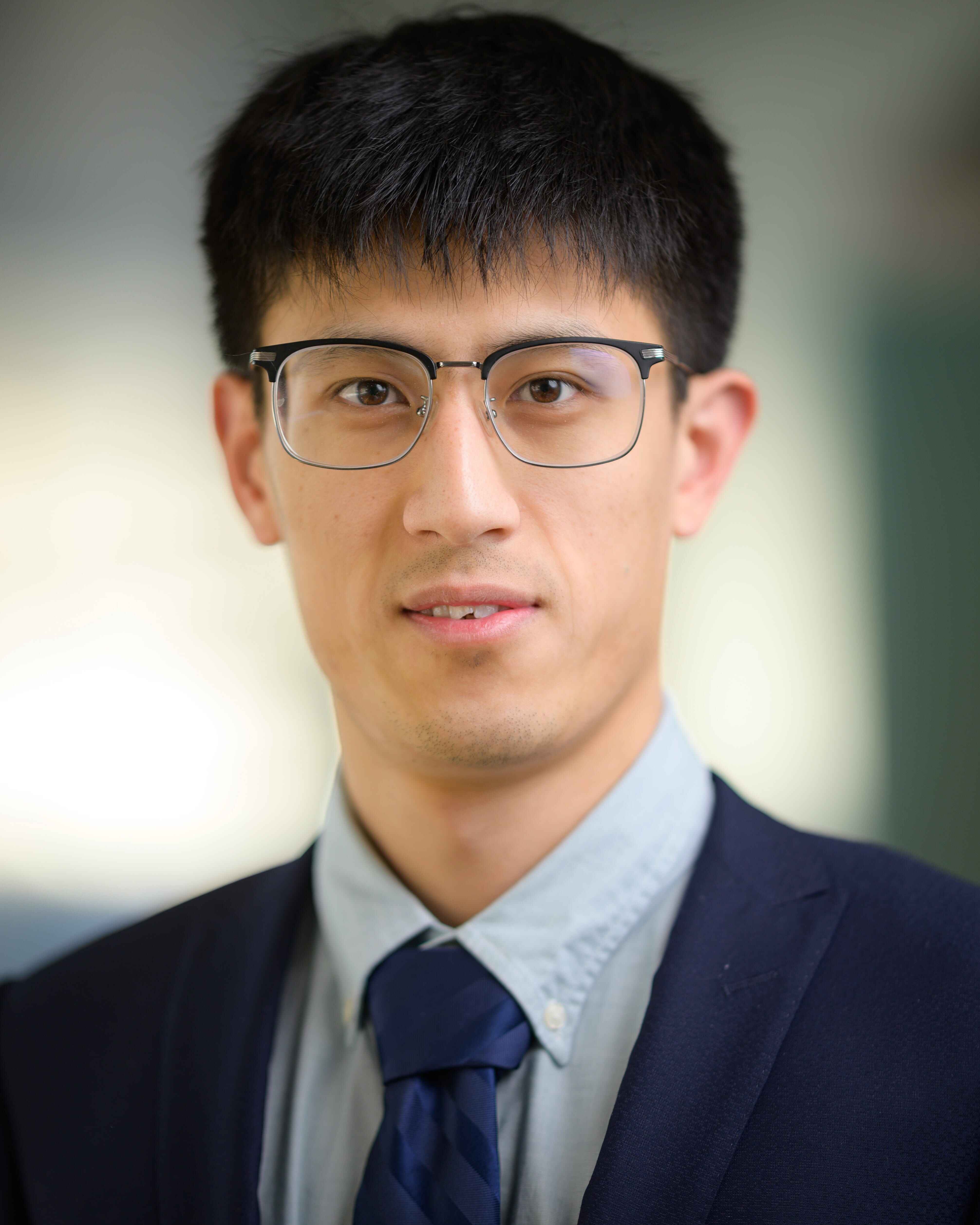}}]{Kaiwen Liu}
received the B.S. degree in electrical and computer engineering from Shanghai Jiao Tong Univeristy, Shanghai, China in 2017, and the M.S. degrees in mechanical engineering and the Ph.D. degree in aerospace engineering from Univeristy of Michigan, Ann Arbor, MI, USA in 2019 and 2023, respectively. His research interests include decision-making with human interactions and learning methods for constrained control problems.
\end{IEEEbiography}

\vspace{-4em}
\begin{IEEEbiography}[{\includegraphics[width=1in,height=1.25in,clip,keepaspectratio]{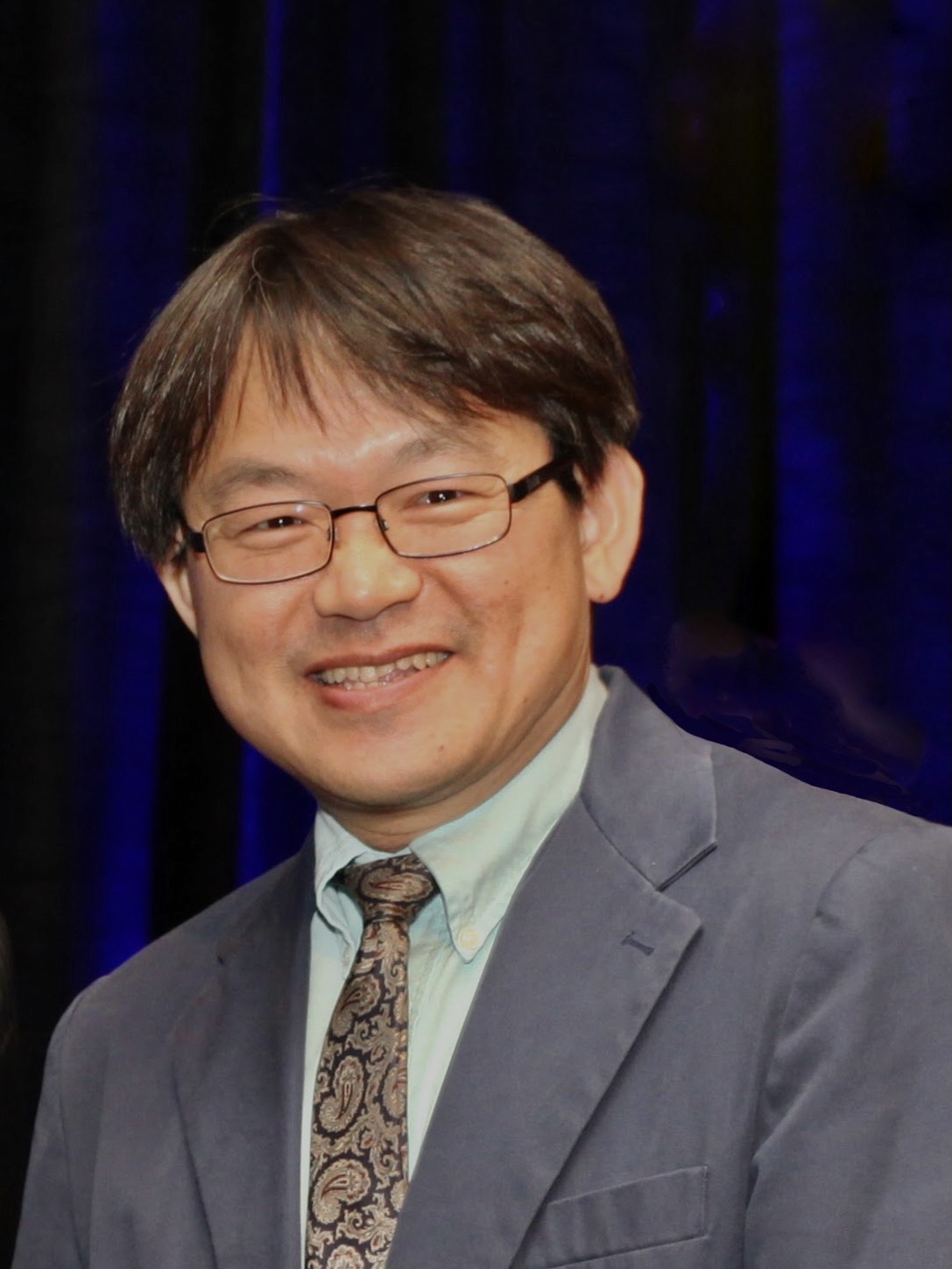}}]{H. Eric Tseng} received the B.S. degree from the National Taiwan University in 1986, and the M.S. and Ph.D. degrees in mechanical engineering from the University of California, Berkeley in 1991 and 1994, respectively. In 1994, he joined Ford Motor Company. At Ford (1994-2022), he had a productive career and retired as a Senior Technical Leader of Controls and Automated Systems in Research and Advanced Engineering. Many of his contributed technologies led to production vehicles implementation. His technical achievements have been honored with Ford's annual technology award, the Henry Ford Technology Award, on seven occasions. Additionally, he was the recipient of the Control Engineering Practice Award from the American Automatic Control Council in 2013. He has over 100 U.S. patents and over 160 publications. He is a member of the National Academy of Engineering as of 2021.
\end{IEEEbiography}

\vspace{-4em}
\begin{IEEEbiography}[{\includegraphics[width=1in,height=1.25in,clip,keepaspectratio]{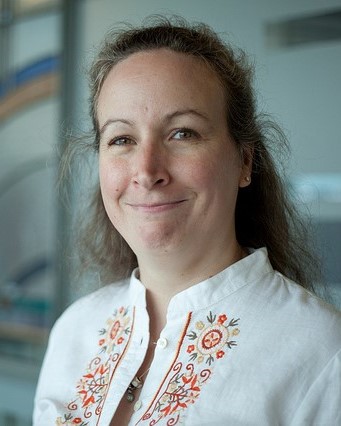}}]{Anouck Girard}
received the Ph.D. degree in ocean engineering from the University of California, Berkeley, CA, USA, in 2002. She has been with the University of Michigan, Ann Arbor, MI, USA, since 2006, where she is currently a Professor of Robotics and Aerospace Engineering, and the Director of the Robotics Institute. Her current research interests include vehicle dynamics and control, as well as decision systems. Dr. Girard was a Fulbright Scholar in the Dynamic Systems and Simulation Laboratory at the Technical University of Crete in 2022.
\end{IEEEbiography}

\vspace{-4em}
\begin{IEEEbiography}[{\includegraphics[width=1in,height=1.25in,clip,keepaspectratio]{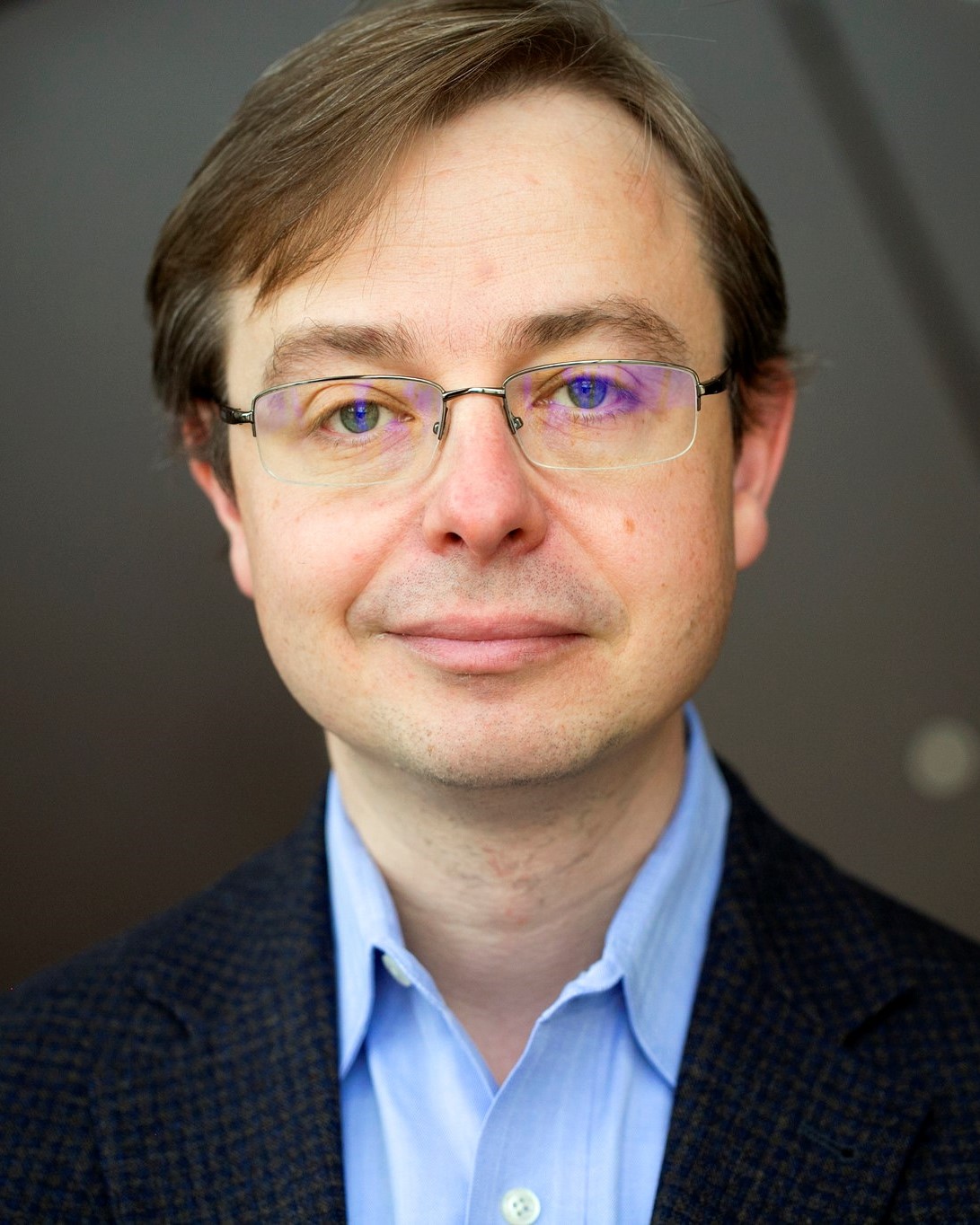}}]{Ilya Kolmanovsky}
received his Ph.D. degree in Aerospace Engineering in 1995 from the University of Michigan, Ann Arbor. He is presently a Pierre T. Kabamba Collegiate Professor in the Department of Aerospace Engineering at the University of Michigan.  His research interests are in control theory for systems with state and control constraints and in aerospace and automotive control applications.
\end{IEEEbiography}

\end{document}